\documentclass[pdflatex,sn-nature]{sn-jnl}

\usepackage{threeparttable}%
\usepackage{graphicx}%
\usepackage{multirow}%
\usepackage{amsmath,amssymb,amsfonts}%
\usepackage{amsthm}%
\usepackage{makecell}
\usepackage{mathrsfs}%
\usepackage[title]{appendix}%
\usepackage{xcolor}%
\usepackage{textcomp}%
\usepackage{manyfoot}%
\usepackage{booktabs}%
\usepackage{algorithm}%
\usepackage{placeins}%
\usepackage{algorithmicx}%
\usepackage{algpseudocode}%
\usepackage{listings}%
\usepackage{caption}
\usepackage{float}%
\usepackage{hhline}
\usepackage{bm}
\usepackage{array}
\usepackage{url}
\usepackage{subcaption}

\theoremstyle{thmstyleone}%
\theoremstyle{thmstyletwo}%

\theoremstyle{thmstylethree}%

\begin{document}

\title[Article Title]{PolyFusionAgent: A Multimodal Foundation Model and Autonomous AI Assistant for Polymer Property Prediction and Inverse Design}






\author{%
\parbox{\textwidth}{\centering
Manpreet Kaur\textsuperscript{a}, Xingying Zhang\textsuperscript{b}, Qian Liu\textsuperscript{a,*} \\[6pt]
\textsuperscript{a}Department of Applied Computer Science, The University of Winnipeg, 515 Portage Avenue, Winnipeg, MB R3B 2E9, Canada. \\
\textsuperscript{b}Department of Mechanical Engineering, University of Manitoba, Winnipeg, MB R3T 2N2, Canada. \\
$^{*}$Corresponding author. E-mail address: qi.liu@uwinnipeg.ca (Q. Liu).
}
}

\abstract{
Polymer discovery is central to fields ranging from energy storage to biomedicine, but it is hindered by an astronomically large chemical design space and fragmented representations of structure, properties, and prior knowledge. This fragmentation leaves many AI models disconnected from physical and experimental reality, restricting their ability to support directly actionable design decisions. Here we introduce PolyFusionAgent, an interactive framework coupling a multimodal polymer foundation model (PolyFusion) with a tool-augmented, literature-grounded design agent (PolyAgent). PolyFusion aligns complementary polymer views—sequence, topology, 3D geometry, and fingerprints—across millions of polymers to learn a shared latent space transferable across chemistries and data regimes, improving thermophysical property prediction and enabling property-conditioned generation of chemically valid, structurally novel polymers beyond the reference design space. PolyAgent closes the design loop by linking prediction and inverse design with evidence retrieval from the polymer literature, proposing, evaluating, and contextualizing hypotheses with explicit precedent in one workflow. Together, PolyFusionAgent enables interactive, evidence-linked polymer discovery combining large-scale representation learning, multimodal chemical knowledge, and verifiable scientific reasoning.
}



\maketitle

\section{Introduction}
Polymers are ubiquitous in modern technology, serving as the backbone of applications from lightweight aerospace composites to flexible electronics, energy storage, catalysis, and biomedicine \cite{mohanty2022sustainable, kesarwani2017polymer}. This versatility arises because polymer properties can be fine-tuned by adjusting monomer sequence, architecture (topology), and processing history \cite{matyjaszewski2005macromolecular, destefano2021biology}. Yet this flexibility is difficult to exploit systematically because polymer representations---and the evidence needed to justify designs---remain fragmented across structure, properties, and prior knowledge. As a consequence, even defining which regions of polymer space are promising is nontrivial; this challenge is compounded by the sheer size of the search space. The polymer chemical space is astronomically large (on the order of $10^{60}$ possible structures) \cite{bohacek1996structure,reymond2015chemical}, and polymer development therefore still relies heavily on slow, empirical cycles of synthesis and testing.  

Early work in polymer informatics \cite{audus2017polymer} applied quantitative structure--property relationships (QSPRs) and machine learning (ML) to predict polymer properties using hand-crafted descriptors such as group contributions and topological indices \cite{vankrevelen2009properties,bicerano2002prediction,joback1987estimation,fedors1974method}. These approaches improved screening efficiency and provided valuable insights, but their fixed, human-designed features limited expressivity and transfer, particularly when extrapolating beyond the chemistries represented in curated datasets. In practice, polymer performance is strongly influenced by factors that are only weakly encoded by idealized repeat-unit formulas or simple descriptors, including chemistry shift, molecular-weight distributions, and processing- or conformation-dependent effects. As a result, conventional polymer ML models often fail under chemical and data-regime shift and struggle to transfer reliably across polymer families and real-world design scenarios. The field therefore needs polymer representations that are both transferable under chemical and data-regime shift and sufficiently rich to support actionable design rather than offline screening.

Foundation models (FMs) (see Supplementary Note 1) trained on massive unlabeled corpora have begun to replace brittle descriptors with learned embeddings for polymer property prediction \cite{kuenneth2023polybert,xu2023transpolymer}. Polymer language models and graph-based encoders capture patterns beyond hand-crafted features and improve screening throughput \cite{rong2020self,wang2022molecular,ying2021transformers}. However, polymer behaviour is governed by information that is only weakly expressed in 1D or 2D encodings, including conformational ensembles, stereoregularity, molecular-weight distributions, end-group chemistry and processing history \cite{sun2025foundation}. Multimodal learning offers a principled route to integrate complementary polymer views, yet existing polymer multimodal models remain comparatively limited in scale and are typically optimized for forward prediction rather than end-to-end design \cite{liu2023multimodal,su2022molecular,huang2025unified,wang2024mmpolymer}. Recent growth in polymer corpora and multimodal molecular representations now makes it feasible to pretrain polymer-specific FMs at scale and to couple them to retrieval and verification tools, enabling design workflows that can be both transferable and evidence-grounded (see Supplementary Note 2).

Beyond forward prediction, a central goal of polymer informatics is inverse design (see Supplementary Note 3): generating new polymer structures that meet specified target properties \cite{savit2025polybart,qiu2024ondemand,gurnani2021polyg2g}. While recent generative models can propose candidates with desired profiles \cite{yue2025benchmarking}, inverse design is only useful in practice if it is built on representations that transfer under chemistry shift and can be linked to supporting evidence. Most polymer generative models are trained independently of FMs and therefore do not benefit from large-scale pretraining, leading to proposals that are often not robust under chemistry shift and lack evidence-grounded justification \cite{struble2024prospective,papidocha2026elephant}. More fundamentally, many design pipelines operate in isolation from the broader scientific context in which polymer decisions are made, limiting their trustworthiness and experimental utility \cite{zhao2025polymer,knox2022autonomous}.

Even with improved representations, inverse design remains non-actionable unless proposals can be checked against prior evidence and constraints. Polymer discovery is inherently knowledge-driven and iterative: researchers consult literature, databases, and prior experiments to assess precedent, feasibility, and trade-offs. However, most polymer AI models remain static offline predictors or generators that lack connectivity to external knowledge and cannot readily verify whether a suggested polymer has precedent, retrieve relevant sources, or incorporate updated findings. Tool-augmented, retrieval-enabled agents in other chemistry domains, such as drug discoveries, illustrate how external knowledge can improve factuality and decision-making \cite{bran2024augmenting,zhong2025benchmarking}. A polymer-specific framework that unifies multimodal polymer representations with retrieval-grounded design has not yet been established.

\begin{figure}[htbp]
    \centering
    \includegraphics[width=0.9\textwidth,keepaspectratio]{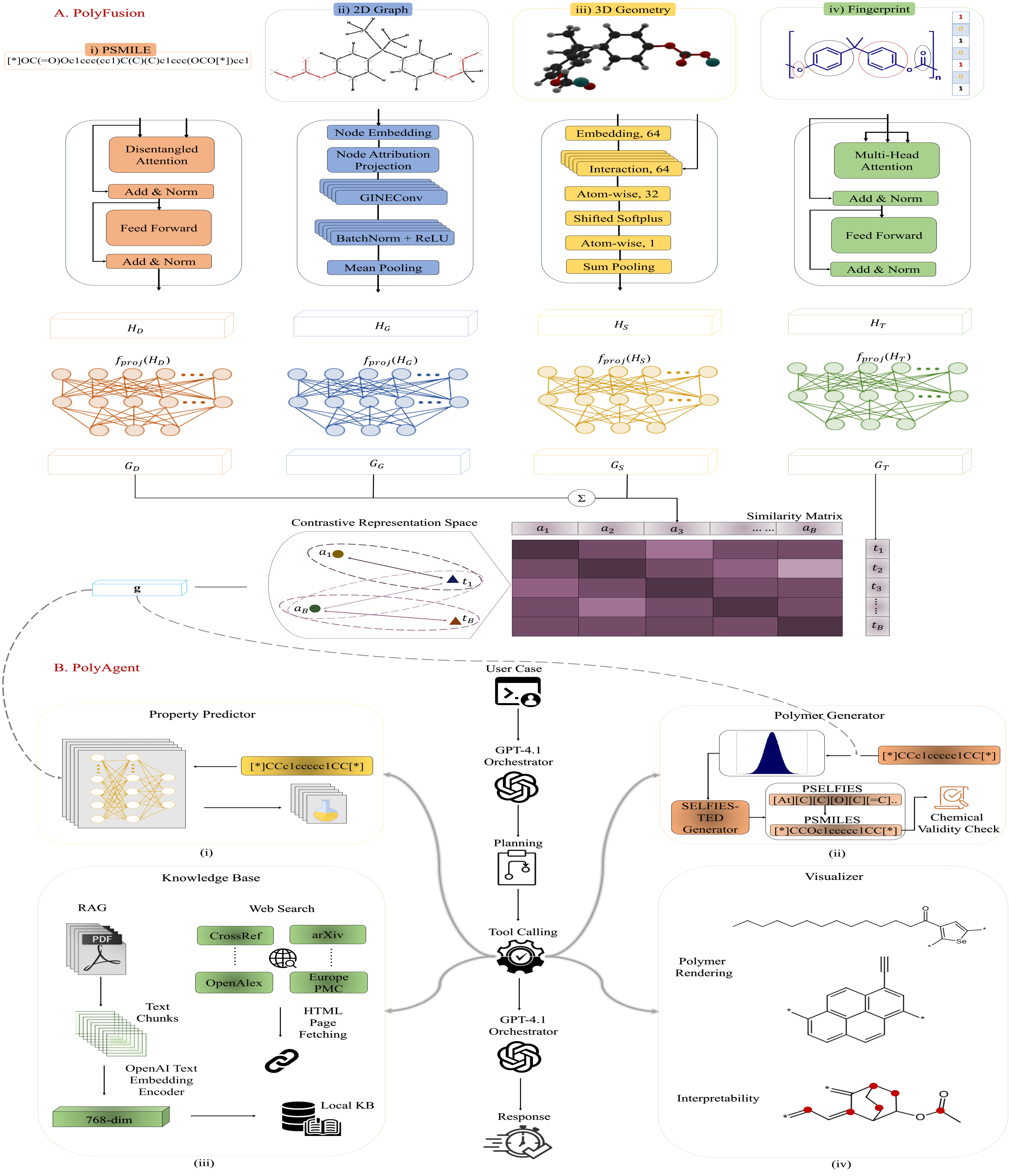}
    \caption{\textbf{Overall framework of the proposed PolyFusionAgent.} It consists of two modules: \textbf{A. PolyFusion}: a multimodal foundation model (FM) and \textbf{B. PolyAgent}: a tool-augmented polymer design AI agent.
    A polymer is represented in four views---(i) PSMILES, (ii)  2D molecular graph, (iii) 3D geometry, and (iv) engineered fingerprint vectors---each mapped to a modality-specific embeddings $H_{\ast}$ by a dedicated encoder: DeBERTaV2 \cite{he2021deberta} for PSMILES, GINE \cite{hu2020strategies} for the 2D graph, SchNet \cite{schutt2018schnet} for 3D structure, and an attention-based Transformer \cite{vaswani2017attention} encoder for fingerprints. A projection head $f_{\mathrm{proj}}(\cdot)$ maps each $H_{\ast}$ into a shared embedding space $G_{\ast}$, where fused structural views yield anchor embeddings $\{a_i\}_{i=1}^{B}$ that are contrastively aligned with fingerprint embeddings $\{t_j\}_{j=1}^{B}$ by computing a cross-similarity matrix $S \in \mathbb{R}^{B \times B}$ and optimizing an InfoNCE loss that treats each matched polymer pair as a positive and all other batch pairs as negatives, maximizing positive similarity while minimizing negative similarity to learn a transferable, cross-view-consistent latent space. For the PolyAgent, a GPT-4.1 orchestrator converts a user request into a plan, performs tool calling, and returns a grounded response. It supports (i) a property predictor for forward inference (structure $\rightarrow$ properties), (ii) a polymer generator for inverse design that outputs candidate polymers in robust string forms (PSMILES) and applies chemical validity checks, (iii) a knowledge base combining retrieval-augmented generation (RAG) over local documents (PDF $\rightarrow$ text chunks $\rightarrow$ embedding encoder $\rightarrow$ vector store) with web search, and (iv) a visualizer that renders polymer structures and exposes interpretability.}
    \label{fig:PolyFusionAgent}
\end{figure}

In this work, we introduce PolyFusionAgent, a unified framework that addresses two coupled barriers in polymer informatics: (i) fragmented polymer representations that limit transfer under chemical and data-regime shift, and (ii) the absence of evidence-grounded workflows that make inverse-design proposals actionable. PolyFusion is a multimodal polymer FM pretrained on multi-million--polymer corpora that aligns complementary polymer views---including Polymer Simplified Molecular Input Line Entry System(PSMILES), 2D graphs, 3D conformational proxies and engineered fingerprints---into a shared latent space that supports both prediction and generation. Building on this backbone, PolyAgent couples model inference with retrieval from the polymer literature and databases to contextualize candidates with prior evidence and constraints, enabling hypotheses to be proposed, evaluated and refined within a single workflow (Fig.\ref{fig:PolyFusionAgent}). Together, PolyFusionAgent transforms polymer informatics from offline screening into an interactive, evidence-linked discovery paradigm.

\section{Experimental Procedure}
\label{sec:methods}

\subsection{Data Curation and Preparation}
\label{sec:data_curation}

We constructed large-scale multimodal training corpora by combining two complementary polymer databases. The pretraining datasets comprised \(2\times 10^{6}\) and \(5\times 10^{6}\) polymers drawn from the union of \texttt{PI1M} (a curated benchmark of hypothetical homopolymers with computed physicochemical properties) and \texttt{polyOne} (an enumerated library of approximately \(10^{8}\) synthetic polymers with associated property annotations). For downstream task evaluation, we used the \texttt{PolyInfo} dataset, a curated collection of approximately \(1.8\times 10^{4}\) experimentally validated polymers with measured and/or literature-compiled properties (see Supplementary Note 4).

\subsection{Multimodal Polymer Representations}
\label{sec:representations}

Each polymer was represented across four complementary modalities. Following Fig.~1, we denote these modalities by \(\ast\in\{D,G,S,T\}\), corresponding to: \(D\) (PSMILES; DeBERTaV2 encoder), \(G\) (2D molecular graph; GINE encoder), \(S\) (3D geometry; SchNet encoder), and \(T\) (engineered fingerprints; Transformer encoder) (see Supplementary Note 5). For RDKit-derived modalities, we converted PSMILES into a chemically valid pseudo-molecule by replacing repeat-unit attachment-point wildcards \texttt{[*]} (a dummy atom) with \texttt{[At]} (Astatine) \cite{mohanty2022sustainable, sahu2026polyt5}, which is rarely present in polymer datasets and serves as an inert terminus marker compatible with standard cheminformatics toolchains.

\subsubsection{Polymer SMILES (PSMILES) sequences (\(D\))}
\label{sec:psmiles}

We represented polymer repeat units as PSMILES strings---text-based molecular descriptors that encode monomer connectivity, stereochemistry, and branching using standard SMILES conventions extended with attachment-point wildcard tokens (denoted \texttt{[*]}) to demarcate repeat-unit boundaries.

PSMILES strings were tokenized using a SentencePiece-based \cite{kudo2018sentencepiece} subword tokenization scheme trained on the pretraining corpora with a fixed vocabulary of 265 tokens. The vocabulary encompasses element symbols (H, He, \(\ldots\), Og), bond symbols (\(-\), \(=\), \(\#\), \(:\)), aromatic markers, and polymer-specific tokens including the \texttt{[*]} attachment-point wildcard. To maintain consistency with learned pretraining embeddings, the vocabulary remained fixed during all downstream fine-tuning tasks.

\subsubsection{2D molecular graphs (\(G\))}
\label{sec:2d_graphs}

Each polymer repeat unit was represented as a labeled molecular graph \(\mathcal{G}=(V,E,X,E_{\mathrm{feat}})\), where \(V\) denotes the set of atomic nodes, \(E\) denotes the set of covalent bond edges, \(X\) denotes node feature attributes, and \(E_{\mathrm{feat}}\) denotes edge feature attributes.

For each atom \(v\in V\), node features comprised: (i) atomic number \(Z_v\in\mathbb{Z}^+\), (ii) formal charge \(q_v\in\mathbb{Z}\), and (iii) a chirality descriptor \(\chi_v\in\{0,1\}^2\) encoding stereochemical configuration.

For each bond \((u,v)\in E\), edge features encoded: (i) bond type \(b_{uv}\in\{1,2,3,4\}\) representing single, double, triple, and aromatic bonds respectively, (ii) stereochemistry \(s_{uv}\in\mathbb{R}\), and (iii) conjugation status \(\kappa_{uv}\in\{0,1\}\) indicating whether the bond participates in a conjugated system.

Graph-structured data were stored in a tensorized format comprising: (i) an adjacency representation encoding bond connectivity and bond types, and (ii) aligned node and edge feature matrices capturing atomic and bonding attributes.

\subsubsection{3D molecular conformations (\(S\))}
\label{sec:3d_conformations}

To capture spatial geometric information beyond 2D topology, we generated three-dimensional atomic coordinates for each polymer repeat unit. Starting from a sanitized RDKit molecular object, we employed the Experimental-Torsion Knowledge Distance Geometry algorithm (ETKDGv3) \cite{riniker2015better} to perform stochastic conformer embedding, followed by Universal Force Field (UFF) \cite{rappe1992uff} optimization to refine local stereochemistry and bond geometry.

The resulting Cartesian coordinate set \(\mathbf{R}=\{r_i\in\mathbb{R}^3 : i=1,\ldots,|V|\}\) defines the spatial positions of atoms corresponding to nodes in \(V\). Each atom \(v_i\) is therefore associated with both a feature vector \(x_i \in \mathbb{R}^{d_x}\) (derived from atomic number and chirality) and a position vector \(r_i \in \mathbb{R}^3\), where \(d_x\) denotes the dimensionality of the atom feature vector. From these coordinates, we compute pairwise Euclidean distances \(D_{ij}=\lVert r_i-r_j\rVert_2\) and bond-angle tensors, which serve as geometric priors for 3D-aware neural network layers.

\subsubsection{Extended-connectivity molecular fingerprints (\(T\))}
\label{sec:fingerprints}

To obtain a compact, fixed-length vector representation suitable for contrastive learning, we computed extended-connectivity fingerprints (ECFP) from the same RDKit-derived molecular topology. Specifically, we used circular fingerprints of radius \(r=3\) (ECFP6) with a bit length of 2048. Each bit in the resulting binary vector encodes the presence or absence of a distinct substructural environment, identified through iterative hashing of atomic neighborhoods up to three bonds away.

The resulting fingerprint vector is denoted \(\mathbf{f}\in\{0,1\}^{2048}\), constituting a high-dimensional, sparsely populated representation that implicitly captures molecular fragments, connectivity patterns, and local chemical motifs.

\subsection{PolyFusion: multimodal contrastive pretraining}
\label{sec:polyfusion}

\subsubsection{Encoders and shared embedding space}
\label{sec:encoders}

PolyFusion comprises four modality-specific encoders: a DeBERTaV2-style Transformer for PSMILES (\(D\)), a GINE message-passing network for 2D graphs (\(G\)), a SchNet continuous-filter network for 3D conformers (\(S\)), and a Transformer encoder for fingerprint bit sequences (\(T\)). For a polymer instance, each encoder produces a pooled, modality-specific hidden representation \(H_{\ast}\in\mathbb{R}^{d_h}\) with \(\ast\in\{D,G,S,T\}\). Subsequently, we map each \(H_{\ast}\) into a shared embedding space via a learnable projection head \(f_{\mathrm{proj}}(\cdot)\), yielding a modality-specific embedding \(G_{\ast}\in\mathbb{R}^{d}\) with \(d=600\):
\begin{equation}
G_{\ast} = f_{\mathrm{proj}}\!\left(H_{\ast}\right)\in \mathbb{R}^{d}
\label{eq:proj_shared}
\end{equation}
We then apply \(\ell_2\)-normalization to place all shared-space embeddings on the unit sphere,
\begin{equation}
G_{\ast} \leftarrow \frac{G_{\ast}}{\left\lVert G_{\ast}\right\rVert_2}
\label{eq:l2_norm_G}
\end{equation}

\subsubsection{Unified masking and corruption}
\label{sec:masking}

Moreover, to regularize pretraining and enable modality-specific reconstruction, we apply a unified masked-unit corruption protocol across modalities. For each modality input \(x^{\ast}\), we sample a masked index set \(\mathcal{M}^{\ast}\) by independently masking units with probability \(p_{\mathrm{mask}}=0.15\). For masked positions, we use an 80--10--10 rule: 80\% are replaced by a modality-specific \texttt{[MASK]} symbol (or masked coordinate token), 10\% are replaced by a random valid symbol/value from the same modality, and 10\% are left unchanged. Denoting the corrupted input by \(\tilde{x}^{\ast}\), the masked reconstruction loss for modality \(\ast\) is
\begin{equation}
\mathcal{L}_{\mathrm{MLM}}^{\ast} =
-\frac{1}{|\mathcal{M}^{\ast}|}
\sum_{i\in \mathcal{M}^{\ast}} \log p_\theta\!\left(x^{\ast}_i \mid \tilde{x}^{\ast}\right)
\label{eq:mlm_generic}
\end{equation}
where the predictive distribution \(p_\theta(\cdot)\) is implemented by a modality-specific reconstruction head attached to the corresponding encoder output.

\subsubsection{Anchor--target multimodal contrastive objective}
\label{sec:contrastive}

PolyFusion, specifically, optimizes an anchor--target InfoNCE objective in which engineered fingerprints (\(T\)) are explicitly designated as the target modality, while the remaining three modalities (\(D\), \(G\), and \(S\)) are aggregated to form a fused structural anchor in the shared embedding space. For a minibatch of \(B\) polymers, let \(G_T^{i}\in\mathbb{R}^{d}\) denote the fingerprint embedding of polymer \(i\), and let \(G_D^{i}\), \(G_G^{i}\), and \(G_S^{i}\) denote its embeddings from the three structural views. We define the fused anchor \(a^{i}\) and the target \(t^{i}\) as
\begin{equation}
a^{i} = \frac{1}{3}\left(G_D^{i} + G_G^{i} + G_S^{i}\right),
\qquad
t^{i} = G_T^{i}
\label{eq:anchor_target_defs}
\end{equation}
We also re-normalize the fused anchor to keep anchors and targets comparable on the unit sphere,
\begin{equation}
a^{i} \leftarrow \frac{a^{i}}{\|a^{i}\|_2}
\label{eq:anchor_renorm}
\end{equation}
Consequently, we form the cross-similarity matrix \(S\in\mathbb{R}^{B\times B}\),
\begin{equation}
S_{ij} = \frac{(a^{i})^{\top} t^{j}}{\tau}
\label{eq:sim_matrix}
\end{equation}
where \(\tau\) is a temperature hyperparameter (\(\tau=0.07\)). The anchor--target InfoNCE loss is
\begin{equation}
\mathcal{L}_{\mathrm{contrast}} =
-\frac{1}{B}\sum_{i=1}^{B}
\log
\frac{\exp(S_{ii})}{\sum_{j=1}^{B}\exp(S_{ij})}
\label{eq:infonce}
\end{equation}
This objective increases similarity for matched polymer pairs \((a_i,t_i)\) while decreasing similarity between each anchor and all non-matching targets in the minibatch.

\subsubsection{Auxiliary reconstruction losses and total objective}
\label{sec:recon_total}

To preserve modality-specific information during alignment, we supplement contrastive learning with masked reconstruction objectives. For structure-native encoders (\(G\) and \(S\)), we use masked atom prediction over atomic numbers. Let \(\mathcal{M}_V\) denote the set of masked atoms and let \(h_v\) denote the final node representation for atom \(v\). The node reconstruction loss is
\begin{equation}
\mathcal{L}_{\mathrm{node}} =
-\frac{1}{|\mathcal{M}_V|}
\sum_{v\in\mathcal{M}_V}\log p_\theta(Z_v\mid h_v)
\label{eq:node_recon}
\end{equation}
For \(D\) (PSMILES) and \(T\) (fingerprints), we use masked token/bit prediction as in Eq.~\eqref{eq:mlm_generic}. We aggregate reconstruction losses across active modalities,
\begin{equation}
\mathcal{L}_{\mathrm{recon}}=
\frac{1}{|\mathcal{S}|}\sum_{\ast\in\mathcal{S}}\mathcal{L}_{\mathrm{MLM}}^{\ast}
\label{eq:recon_mean}
\end{equation}
where \(\mathcal{S}\subseteq\{D,G,S,T\}\) is the set of modalities with an active reconstruction head for the current batch. The total pretraining objective is
\begin{equation}
\mathcal{L}_{\mathrm{total}} = \mathcal{L}_{\mathrm{contrast}} + \lambda\,\mathcal{L}_{\mathrm{recon}},
\qquad \lambda = 1.0
\label{eq:total_loss}
\end{equation}

\subsubsection{Training configuration}
\label{sec:training_config}

We trained PolyFusion\_2M and PolyFusion\_5M with AdamW (learning rate \(\alpha=10^{-4}\), weight decay \(10^{-2}\)) using a base batch size \(B_{\mathrm{base}}=16\) and gradient accumulation over 4 steps (effective batch size \(B_{\mathrm{eff}}=64\)). Models were trained for up to 25 epochs with early stopping (patience 10) on the validation objective \(\mathcal{L}_{\mathrm{total}}\). Mixed-precision training (FP16) was used to reduce memory footprint and improve throughput, and all pretraining runs were executed on NVIDIA RTX A6000 GPUs (48~GB VRAM). To stabilize multimodal optimization, each modality encoder was first initialized via unimodal masked reconstruction pretraining (Eq.~\eqref{eq:mlm_generic}) before joint anchor--target contrastive training.

\subsection{Downstream adaptation}
\label{sec:downstream}

\subsubsection{Thermophysical property prediction}
\label{sec:downstream_prediction}
Thermophysical properties are predicted using a lightweight regression head trained on top of the frozen encoders. Specifically, we use a two-layer MLP regressor with a hidden width of \(300\) and a ReLU nonlinearity; this regressor takes as input the \(\ell_2\)-normalized fused polymer embedding \(\mathbf{g}\) in \(\mathbb{R}^{600}\) and outputs a scalar prediction \(\hat{y}\), with dropout applied to the hidden activations at probability \(p_{\mathrm{drop}}=0.1\). During this downstream training, only the parameters of the projection layers and the regression head are optimized, while all pretrained encoder parameters remain fixed.

We optimize the model using MSE on standardized targets:
\begin{equation}
\mathcal{L}_{\mathrm{MSE}}(y,\hat{y})=(y-\hat{y})^2
\label{eq:mse}
\end{equation}
where \(y\) denote the scaled target. The target standardization is performed with a \texttt{StandardScaler} fit on the training split only, and predictions are inverse-transformed to the original units for reporting.

For each property, we evaluate performance using fivefold cross-validation (\(K=5\)), and within each fold we reserve an inner validation split from the training fold (10\% of the train-val subset) for early stopping and model selection. We use Adam optimization with learning rate \(\alpha=10^{-4}\) and weight decay \(0.0\), together with a cosine annealing learning-rate schedule over up to 100 epochs. Early stopping (patience 10) is applied based on validation MSE, and we report mean \(\pm\) s.d.\ over the five folds. On the held-out test split for each run, we evaluate performance using the coefficient of determination (\(R^2\)), which measures the fraction of variance in the targets explained by the predictions, along with mean absolute error (MAE) and root-mean-squared error (RMSE), which summarize absolute and squared prediction deviations, respectively.

\subsubsection{Property-conditioned polymer generation}
\label{sec:generation}
For inverse design we use a property-conditioned sequence-to-sequence generator in which PolyFusion provides a fixed embedding space for conditioning and retrieval. Given a polymer, PolyFusion produces a unit-norm latent embedding \(\mathbf{g}\in\mathbb{R}^{600}\), \(\|\mathbf{g}\|_2=1\). A pretrained SELFIES-based encoder--decoder model (SELFIES-TED) is fine-tuned to reconstruct polymer PSELFIES strings from \(\mathbf{g}\) by mapping it into a short set of \(K=4\) learned memory tokens that serve as the encoder-side states:
\begin{equation}
\mathbf{m} = W_{\mathrm{cond}} \mathbf{g}\in\mathbb{R}^{d_{\mathrm{model}}},
\qquad
\mathbf{M}_k = \mathbf{m} + \mathbf{p}_k,\;\; k=1,\ldots,K,
\qquad
\mathbf{M}\in\mathbb{R}^{K\times d_{\mathrm{model}}}
\label{eq:mem_tokens}
\end{equation}
where $d_{\mathrm{model}}$ is the hidden dimension of the SELFIES-TED encoder--decoder, \(W_{\mathrm{cond}}\) is a learned mapping implemented as a small feed-forward projection, and \(\{\mathbf{p}_k\}\) are learned positional embeddings for the memory slots.

During generator training, we inject Gaussian noise into the latent and re-normalize to encourage local robustness,
\begin{equation}
\tilde{\mathbf{g}} = \frac{\mathbf{g}+\boldsymbol{\eta}}{\|\mathbf{g}+\boldsymbol{\eta}\|_2},
\qquad
\boldsymbol{\eta}\sim\mathcal{N}(0,\sigma_{\mathrm{train}}^{2}I),
\qquad
\sigma_{\mathrm{train}}=0.10
\label{eq:latent_noise}
\end{equation}
and compute \(\mathbf{M}\) from \(\tilde{\mathbf{g}}\). The generator is trained with teacher forcing to maximize the conditional likelihood of the ground-truth PSELFIES sequence \((s_1^\ast,\ldots,s_T^\ast)\),
\begin{equation}
\mathcal{L}_{\mathrm{gen}} = -\frac{1}{T}\sum_{t=1}^{T}\log p(s_t^\ast \mid s_{<t}^\ast, \mathbf{M})
\label{eq:lgen}
\end{equation}
At inference time, we decode using top-\(p\) sampling (default \(p=0.92\)) and temperature (default \(1.0\)), with repetition penalty \(1.05\), stopping on \texttt{</s>} or a maximum length of 256 tokens (with a minimum length of 10 tokens). The decoder outputs polymers as PSELFIES strings; for downstream evaluation and compatibility with cheminformatics tooling, we deterministically convert PSELFIES to PSMILES via the standard SELFIES decoding and canonicalization. 

Specifically, property targeting is performed via generate-then-filter in latent space using a Gaussian-process regression oracle \(\hat{g}(\cdot)\) trained on \((\mathbf{g},y)\) pairs for the target property. Therefore, for a scaled target $y_{\mathrm{target}}$, we choose a small set of training seeds whose scaled properties are closest to $y_{\mathrm{target}}$, perturb each seed latent with Gaussian noise $\sigma_{\mathrm{gen}}=0.15$ and re-normalize to unit norm, decode each resulting $\tilde{\mathbf{g}}$ to a candidate polymer, and keep only those candidates satisfying $\left|\hat{g}(\tilde{\mathbf{g}})-y_{\mathrm{target}}\right|\le \tau_s$ with $\tau_s=0.5$ in standardized units. We additionally summarize generation quality using standard validity, novelty, and diversity metrics: validity is the fraction of decoded candidates that yield chemically valid polymers, novelty is the fraction of valid candidates not present in the training set, and diversity is quantified via the average pairwise dissimilarity (computed over the valid set) to assess structural spread among generated polymers. All generation metrics are computed per split within the \(K=5\) cross-validation protocol and then aggregated across folds to report the mean and standard deviation.

\subsection{PolyAgent: tool-mediated polymer discovery}
\label{sec:polyagent}
We implemented PolyAgent as a tool-mediated inference pipeline that couples the pretrained PolyFusion\_5M representation (Section~\ref{sec:polyfusion}) to (i) property prediction and (ii) property-conditioned polymer generation, and integrates additional tools for (iii) retrieval-augmented (local + web) evidence synthesis with explicit citation and (iv) visualization. Specifically, a large language model (GPT-4.1) was used as a controller to translate each user query into a finite sequence of tool calls with explicit data dependencies. The controller was run with temperature \(\tau_{\mathrm{plan}}=0.2\) to reduce stochastic variation in planning.

\subsubsection{Property Predictor}
\label{sec:polyagent_property}
For property inference, we constructed the PolyFusion multimodal inputs from the query PSMILES (Section~\ref{sec:representations}), computed the unit-normalized polymer embedding \(\mathbf{g}\in\mathbb{R}^{600}\) using the pretrained PolyFusion\_5M encoder (Sections~\ref{sec:encoders}--~\ref{sec:contrastive}), and predicted target properties using the downstream regression heads described in Section~\ref{sec:downstream_prediction}. Concretely, for each property \(k\), we applied a two-layer MLP regressor with hidden width 300 and ReLU activation to \(\mathbf{g}\) to obtain \(\hat{y}_k\). Targets and predictions were standardized/inverse-transformed exactly as in Section~\ref{sec:downstream_prediction} to report values in physical units.

\subsubsection{Polymer Generator}
\label{sec:polyagent_generation}
For inverse design, PolyAgent invoked the property-conditioned generator described in Section~\ref{sec:generation}. Given a target property \(k\) and standardized target value \(y_{\mathrm{target}}\), the system generated candidates via latent seeding and perturbation followed by constrained decoding, and retained candidates using the property-oracle acceptance criterion.

\subsubsection{Knowledge Base}
\label{sec:polyagent_rag}
To ground PolyAgent responses in external evidence, we curated a local document corpus and also implemented web augmentation.

\paragraph{Local corpus construction and indexing}
We assembled a local corpus comprising \(N_{\mathrm{docs}}=1108\) PDFs spanning six categories: (i) 10 curated standards (IUPAC nomenclature and ISO/ASTM testing protocols), (ii) 38 open-access journal articles (\textit{Macromolecules}, \textit{Polymer Chemistry}, \textit{ACS Polymers Au}), (iii) 600 arXiv preprints (cond-mat.mtrl-sci, cond-mat.soft, physics.chem-ph, cs.LG, stat.ML), (iv) 400 OpenAlex polymer publications, and (v) 60 Europe PMC biopolymer articles. Documents were then segmented into overlapping passages at three granularities (512/256/128 tokens; overlaps 64/48/32 tokens) using the TikToken \texttt{cl100k\_base} tokenizer, yielding \(N_{\mathrm{chunks}}=6.2\times 10^6\) retrievable segments. Duplicates were removed via SHA256 content hashing. Afterwards, all chunks were embedded using OpenAI \texttt{text-embedding-3-small} (\(d_{\mathrm{emb}}=1536\)) \cite{openai2024textembedding3small} and indexed in FAISS using an HNSW graph (\(M=64\), \(ef_{\mathrm{construction}}=200\)) \cite{malkov2020hnsw}. At query time, for a query \(\mathbf{q}\), the top-\(k\) chunks were retrieved by cosine similarity in the embedding space and reranked using a cross-encoder (\texttt{ms-marco-MiniLM-L-12-v2}); the top-\(k_{\mathrm{rerank}}=5\) passages were selected to construct the evidence context \(\mathbf{C}\).  

\paragraph{Web Search Augmentation}
For web search, seven sources (CrossRef, arXiv, OpenAlex, Europe PMC, Semantic Scholar, Springer Nature, and Internet Archive Scholar) were queried using a reformulated query produced by a T5-base rewrite model trained on \(8.0\times 10^3\) rewrite pairs to expand abbreviations and add temporal constraints; the final response model was constrained to synthesize answers using only the resulting evidence context \(\mathbf{C}\) and to attach citations for statements derived from retrieved passages (including title, authors, year, and DOI/identifier when available), reporting evidence gaps and avoiding uncited numerical claims when support for a requested quantitative value was absent.
For each source, retrieved items were scored using a weighted relevance--recency--trust function:
\begin{equation}
s_{\mathrm{web},i} = 0.6\,\mathrm{BM25}(\mathbf{q}, \mathrm{snippet}_i) + 0.3\,\exp(-0.5\,\Delta t_i) + 0.1\,\mathbb{I}_{\mathrm{trusted}}(\mathrm{source}_i)
\label{eq:polyagent_webscore}
\end{equation}
where \(\Delta t_i\) is years since publication and \(\mathbb{I}_{\mathrm{trusted}}\) indicates peer-reviewed venues \cite{robertson2009probabilistic}. We fetched the top-3 results per source, extracted text using Trafilatura, chunked and embedded the text as above, and appended these passages to \(\mathbf{C}\).

\subsubsection{Visualizer}
\label{sec:polyagent_viz}
We generated structure visualizations by converting PSMILES to RDKit molecules and rendering 2D depictions using \texttt{MolsToGridImage}. For interpretability of property predictions, we computed atom-level occlusion attributions using a leave-one-atom-out protocol. For an input polymer \(x\) and a selected property predictor \(\hat{y}_k(\cdot)\), we computed a baseline prediction \(y_0=\hat{y}_k(x)\). For each atom \(i\), we formed an occluded structure \(x^{(-i)}\) by replacing atom \(i\) with a wildcard atom (atomic number 0) and re-sanitizing the molecule, then evaluated \(y_i=\hat{y}_k(x^{(-i)})\). We defined the attribution score as
\begin{equation}
s_i = y_0 - y_i
\label{eq:polyagent_occlusion}
\end{equation}
We selected highlighted atoms by thresholding relative to the maximum occlusion magnitude and applying a top-\(K\) cap:
\begin{equation}
|s_i|\ge \max\!\left(\beta,\ \alpha \max_j |s_j|\right),
\qquad \alpha=0.25,\ \beta=0,\ K=12
\label{eq:polyagent_select}
\end{equation}


\subsubsection{Evaluation}
\label{sec:baselines}
To contextualize PolyAgent's performance, we compared it against LLM-only (no tools) baselines. Concretely, we evaluate two non-GPT models exposed in the UI under the same no-tools constraint: Mixtral-8x22B-Instruct (\texttt{mistralai/Mixtral-8x22B-Instruct-v0.1}) and Llama-3.1-8B-Instruct (\texttt{meta-llama/Llama-3.1-8B-Instruct}).

Performance was quantified using five metrics \cite{ferber2025development}, computed over a simulated suite of 20 diverse polymer-design cases. Each metric was scored per case on a 0--10 scale and then aggregated across cases.

\paragraph{Tool Use}
Tool use measures whether required tool calls, when invoked, executed successfully and returned minimally valid outputs. For case \(c\), let \(T_{\mathrm{req}}(c)\) denote the set of required tool calls and let \(T_{\mathrm{succ}}(c) \subseteq T_{\mathrm{req}}(c)\) denote those that executed and returned minimally valid outputs. The per-case tool-use fraction is
\begin{equation}
\text{Tool Use}(c) = \frac{\lvert T_{\mathrm{succ}}(c) \rvert}{\lvert T_{\mathrm{req}}(c) \rvert}
\end{equation}
The integer score on the 0--10 scale is computed as \(\mathrm{round}\!\left(10 \cdot \text{Tool Use}(c)\right)\), where the rounding rule and bounds are defined below.

\paragraph{Completeness}
Completeness measures whether the system invoked all tool steps deemed essential for a given case. Let \(T_{\mathrm{used}}(c)\) be the set of tools actually invoked. We define
\begin{equation}
\text{Completeness}(c) =
\frac{\lvert T_{\mathrm{used}}(c) \cap T_{\mathrm{req}}(c) \rvert}{\lvert T_{\mathrm{req}}(c) \rvert}
\end{equation}
and convert this fraction to an integer 0--10 score by \(\mathrm{round}\!\left(10 \cdot \text{Completeness}(c)\right)\).

\paragraph{Correctness}
Correctness evaluates factual accuracy and logical consistency with respect to (i) the case prompt and (ii) any available tool outputs. Scores are assigned on the 0--10 scale according to predefined anchors: 10 indicates full consistency; 7--8 indicates mostly correct responses with minor inaccuracies; 4--6 indicates mixed quality with partially unsupported claims; and 0--3 indicates substantial hallucination or flawed reasoning. Scores are reported as integers in \([0,10]\); when a fractional assessment is performed, the reported value is obtained using \(\mathrm{round}(\cdot)\) as described below.

\paragraph{Helpfulness}
Helpfulness assesses practical usefulness, clarity, and actionability for polymer scientists. Following \cite{ferber2025development}, helpfulness is computed as the weighted sum
\begin{equation}
\text{Helpfulness}(c)
= 0.4\,\eta_{\mathrm{enum}}
+ 0.3\,\eta_{\mathrm{toolref}}
+ 0.3\,\eta_{\mathrm{format}}
\end{equation}
where \(\eta_{\text{enum}}\) measures whether the response clearly enumerates the requested deliverables (e.g., action items, ranked candidates, or a stepwise screening workflow), \(\eta_{\text{toolref}}\) measures whether tool outputs are explicitly referenced and integrated into the response (e.g., predicted values and/or retrieved document DOIs), and \(\eta_{\text{format}}\) measures clarity and structure (e.g., headings, separation of reasoning vs.\ recommendations, and constraint checklists), with each \(\eta \in [0,1]\). The composite is converted to the 0--10 integer scale via \(\mathrm{round}\!\left(10 \cdot \text{Helpfulness}(c)\right)\). When component scores cannot be reliably enumerated, a holistic integer score is assigned using the same convention, and annotators record the reasons for any deviation.

\paragraph{Citation Accuracy}
Citation accuracy measures whether citations provided in the response resolve to retrievable sources and correspond to the referenced documents. If a response contains \(N_{\mathrm{cit}}(c)\) citations and \(N_{\mathrm{ok}}(c)\) of these are verified as correct, then
\begin{equation}
\text{Citation Accuracy}(c) =
\frac{N_{\mathrm{ok}}(c)}{N_{\mathrm{cit}}(c)}
\end{equation}
The 0--10 integer score is computed as \(\mathrm{round}\!\left(10 \cdot \text{Citation Accuracy}(c)\right)\). For tasks that explicitly request literature support, responses without citations are assigned low default scores (0--2) unless a credible justification for the absence of citations is provided.

\section{Results}
\label{sec:results}

\subsection{Scale and multimodal fusion improve polymer representation learning}
\begin{figure}[htbp]
    \centering
    \includegraphics[width=\textwidth,keepaspectratio]{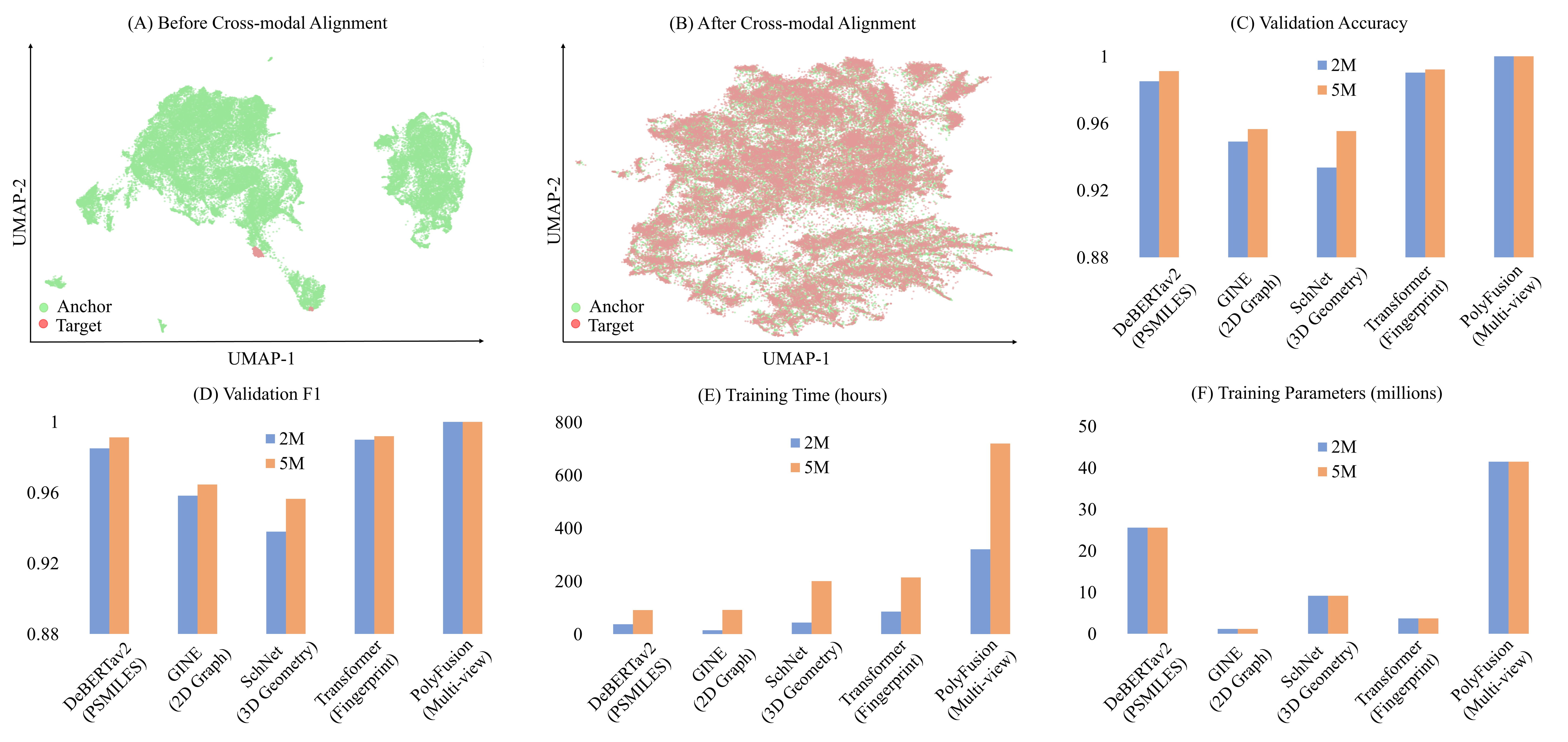}
    \caption{\textbf{Scaling and cross-view alignment strengthen polymer embeddings.} \textbf{(A)} Embedding distributions of PolyFusion\_5M before contrastive anchor–target alignment. Embeddings from the anchor views (PSMILES, 2D graph, and 3D proxy) are shown in green, while embeddings from the target view (fingerprint) are shown in red. The target-view embeddings collapse into a narrow subregion of the anchor embedding space, indicating strong residual modality dependence and limited cross-modal alignment prior to multimodal contrastive training. \textbf{(B)} Corresponding visualization after anchor–target alignment, where anchor and target embeddings co-localize across the dominant manifold, indicating effective cross-modal integration. \textbf{(C)} Validation accuracy, \textbf{(D)} validation F1, \textbf{(E)} training time (hours), and \textbf{(F)} trainable parameters (millions) for unimodal encoders---DeBERTaV2 (PSMILES), GINE (2D graph), SchNet (3D proxy), and a Transformer (fingerprint)---and PolyFusion under 2M and 5M pretraining.}
    \label{fig:scale_fusion}
\end{figure}

PolyFusion is designed around a simple premise: polymer behaviour is governed by coupled chemical and structural factors that are only partially observable in any single representation. PSMILES sequences are information-rich but do not explicitly encode local graph neighbourhood constraints or 3D interaction geometry; 2D graphs resolve local connectivity and functional-group environments yet can underrepresent longer-range context and conformational degrees of freedom; 3D conformational proxies introduce distance-conditioned interaction priors but are intrinsically noisy at repeat-unit scale because they reflect a single embedded conformer rather than an ensemble; and engineered fingerprints compress substructural information into a fixed-length similarity space while inheriting the inductive bias of hashing and radius truncation. PolyFusion therefore treats cross-view consistency as a supervisory signal in pretraining. Multi-view representations (PSMILES, 2D graph, 3D conformational proxies, and fingerprint embedding) of 2 million (2M) and 5 million (5M) polymers are collected from PI1M and polyOne corpora (see Supplementary Note 4). A contrastive learning strategy is then employed to encourage consistency between the anchor view (PSMILES, 2D, and 3D views) and the target view (the fingerprint view), ensuring that the final fused multi-view representation is aligned across modalities. As visualized by the Uniform Manifold Approximation and Projections (UMAPs) in Fig.\ref{fig:scale_fusion},(A) (PolyFusion\_5M), the resulting latent spaces retain measurable modality dependence: the anchor and target embeddings occupy related but not fully co-localized regions, and their manifold structure shows residual view-specific distortion. This behaviour is expected—each modality emphasizes a different set of invariances (syntax-level regularities for sequences, local neighbourhood isomorphisms for graphs, distance-based interaction priors for geometry, and substructure-presence statistics for fingerprints), so independently trained encoders can be individually performant while remaining geometrically misaligned. PolyFusion addresses this fragmentation by explicitly enforcing alignment between modalities via contrastive learning. Fig.\ref{fig:scale_fusion},(B) shows that anchor and target embeddings co-localize across the dominant manifold rather than partitioning into modality-defined clusters. In practice, this indicates that PolyFusion learns a shared coordinate system in which polymers can be retrieved, compared, and conditioned upon consistently regardless of whether they are presented as sequences, graphs, conformers, or fingerprints—providing the representation stability required for downstream transfer and inverse design under realistic heterogeneity in polymer inputs.  

The unimodal encoders (see Supplementary Note 5) are able to learn meaningful polymer representations on their own (Fig.\ref{fig:scale_fusion},(C-F)); however, they reveal a clear hierarchy in how effectively different input modalities support discriminative pretraining. Consistent with their information-rich, highly engineered representations, the sequence encoder (Decoding-Enhanced BERT with Disentangled Attention (DeBERTaV2) \cite{he2021deberta}) and the Morgan fingerprints encoder (Transformer \cite{vaswani2017attention}) rapidly achieve strong validation accuracy and F1 score with 2M pretraining samples (Fig. \ref{fig:scale_fusion}, C–D), and exhibit only marginal gains when scaling to 5M samples, indicating early saturation under the contrastive pretraining objective. By contrast, the structure-native encoders exhibit a much stronger response to increased pretraining scale: both the 2D graph model (Graph Isomorphism Network with Edge features (GINE) \cite{hu2020strategies}) and the 3D proxy model (a continuous-filter convolutional neural network (SchNet) \cite{schutt2018schnet}) show clear gains when scaled to 5M samples, with SchNet exhibiting the largest relative improvement. This pattern is consistent with geometric information being both highly informative and noisier at the repeat-unit level in polymer chains, making 3D representations especially data-hungry to learn stable invariances. Importantly, PolyFusion strengthens performance beyond any single-view encoder at both 2M and 5M scale regimes, with the largest and most consistent gains. This pattern supports the design claim: explicitly aligning the fingerprint target to a fused structural anchor does not merely average complementary signals, but reduces view-specific failure modes and yields a representation whose discriminative structure is more stable than that of its constituents.

From a systems perspective, these gains are not driven by a large increase in model size: PolyFusion has a parameter count in the same order-of-magnitude as the strongest single encoders (Fig. \ref{fig:scale_fusion}, F), indicating that the improvements mainly come from the supervision design (cross-view alignment and fusion) rather than brute-force parameter scaling. The main cost is computational rather than parametric (Section~\ref{sec:training_config}). Training time increases with dataset size for all models, but grows fastest for PolyFusion because each sample is processed by multiple encoders and optimized with cross-view objectives (Fig. \ref{fig:scale_fusion}, E). Importantly, this cost is offset by PolyFusion’s role as a single reusable representation for both prediction and generation: once pretrained, the unified embedding avoids the need to redesign descriptors or train separate modality-specific models for different downstream tasks. Overall, the results highlight a favorable efficiency trade-off for polymer FM pretraining: modest parameter overhead and higher training cost yield a substantially more transferable latent space, especially when scaling from 2M to 5M diverse polymers.

\subsection{Fused embeddings improve thermophysical property prediction}
To test whether the fused representation captures polymer physics beyond the pretraining alignment objective, we evaluated their performance on four thermophysical property predictions: density ($\rho$), glass-transition temperature ($T_g$), melting temperature ($T_m$) and thermal decomposition temperature ($T_d$). These properties span distinct but coupled physical controls: $\rho$ reflects packing and cohesive energy density; $T_g$ probes segmental mobility and free-volume constraints; $T_m$ depends on crystallizability and chain regularity; and $T_d$ reports chemical robustness and the stability of the weakest links under thermal stress. Together, they form a compact benchmark that is both application-relevant and mechanistically diverse.

For downstream property prediction, we adapted PolyFusion (and all baseline models) by adding a small two-layer neural network on top of the learned polymer representation. This network was trained using a robust regression loss (mean-squared error, MSE), while the pretrained representation was kept frozen during training to prevent overfitting and preserve general chemical features (Section~\ref{sec:downstream_prediction}). Across all four properties, PolyFusion exhibits the most reliable transfer behaviour, with PolyFusion\_5M (PolyFusion pretrained on 5M polymers) consistently providing the strongest generalization compared to other baseline models, as reflected in both explained variance and performance metrics (Table \ref{tab:thermo_polyinfo}). The advantage is most pronounced for the transition and stability temperatures ($T_g$, $T_m$, $T_d$), where predictive performance depends on joint sensitivity to local chemistry (e.g., polarity, bond strengths, functional-group environments) and higher-order effects (e.g., conformational rigidity, packing frustration, intermolecular constraints). In these regimes, multimodal fusion behaves as a physically meaningful regularizer: when one view is noisy or underdetermined (for instance, approximate conformers or repeat-unit ambiguities), complementary views preserve property-relevant information in the shared latent geometry.

\begin{table*}[h!]
\centering
\caption{\textbf{Performance on thermophysical property prediction.} Unimodal baselines (ChemBERTa, PolyCL, PolyBERT), multimodal baselines (MMPolymer, MoleculeSTM, PolyNC), and the proposed PolyFusion models pretrained at two scales (PolyFusion\_2M and PolyFusion\_5M) are fine-tuned and evaluated on four representative thermophysical properties: density ($\rho$), glass-transition temperature ($T_g$), melting temperature ($T_m$), and thermal-decomposition temperature ($T_d$). Performance is assessed using the coefficient of determination ($R^2$, higher is better) and prediction error measured by mean absolute error (MAE) and root-mean-square error (RMSE) (lower is better). Results are reported as mean $\pm$ standard deviation over five independent cross-validation folds. Best and second-best values for each metric are highlighted in \textbf{bold} and \underline{underlined}, respectively. Overall, multimodal fusion—particularly PolyFusion\_5M—consistently improves explained variance and reduces prediction error compared with unimodal baselines, with the largest gains observed for thermal transition and stability properties ($T_g$, $T_m$, and $T_d$).
}
\label{tab:thermo_polyinfo}

\resizebox{\linewidth}{!}{%
\begin{tabular}{|l|l|l|c|c|c|}
\hline
\textbf{Property} & \textbf{Input Setting} & \textbf{Model} & $\mathbf{R^2}$ (\(\uparrow\)) & \textbf{MAE} (\(\downarrow\)) & \textbf{RMSE} (\(\downarrow\)) \\
\hline

\multirow{8}{*}{$\rho$}  & \multirow{3}{*}{Unimodal}   & ChemBERTa \cite{chithrananda2020chemberta}   & 0.679 $\pm$ 0.046 & 0.069 $\pm$ 0.002 & 0.115 $\pm$ 0.012 \\
&                             & PolyCL \cite{zhou2025polycl}      & 0.735 $\pm$ 0.061 & 0.061 $\pm$ 0.003 & 0.104 $\pm$ 0.016 \\
&                             & PolyBERT \cite{kuenneth2023polybert}    & 0.717 $\pm$ 0.069 & 0.057 $\pm$ 0.004 & 0.108 $\pm$ 0.017 \\
\hhline{|~|-|-|-|-|-|}
& \multirow{5}{*}{Multimodal} & MMPolymer \cite{wang2024mmpolymer}   & 0.732 $\pm$ 0.056 & 0.062 $\pm$ 0.002 & 0.105 $\pm$ 0.015 \\
&                             & MoleculeSTM \cite{liu2023multimodal} & 0.740 $\pm$ 0.060 & 0.062 $\pm$ 0.003 & 0.103 $\pm$ 0.016 \\
&                             & PolyNC \cite{qiu2024polync}      & \underline{0.768 $\pm$ 0.041} & \underline{0.053 $\pm$ 0.005} & 0.104 $\pm$ 0.013 \\
&                             & PolyFusion\_2M (Ours) & 0.756 $\pm$ 0.061 & 0.053 $\pm$ 0.004 & \underline{0.100 $\pm$ 0.016} \\
&                             & PolyFusion\_5M (Ours) & \textbf{0.776 $\pm$ 0.050} & \textbf{0.052 $\pm$ 0.003} & \textbf{0.095 $\pm$ 0.014} \\
\hline

\multirow{8}{*}{$T_g$}  & \multirow{3}{*}{Unimodal}   & ChemBERTa \cite{chithrananda2020chemberta}   & 0.866 $\pm$ 0.006 & 27.740 $\pm$ 0.992 & 40.440 $\pm$ 1.050 \\
&                             & PolyCL \cite{zhou2025polycl}      & 0.814 $\pm$ 0.009 & 34.620 $\pm$ 0.842 & 47.700 $\pm$ 0.898 \\
&                             & PolyBERT \cite{kuenneth2023polybert}    & 0.882 $\pm$ 0.006 & 26.010 $\pm$ 0.814 & 37.940 $\pm$ 0.903 \\
\hhline{|~|-|-|-|-|-|}
& \multirow{5}{*}{Multimodal} & MMPolymer \cite{wang2024mmpolymer}   & 0.825 $\pm$ 0.017 & 33.090 $\pm$ 2.170 & 46.290 $\pm$ 2.250 \\
&                             & MoleculeSTM \cite{liu2023multimodal} & 0.808 $\pm$ 0.006 & 35.260 $\pm$ 0.949 & 48.450 $\pm$ 0.702 \\
&                             & PolyNC \cite{qiu2024polync}      & \underline{0.900 $\pm$ 0.004} & \underline{23.670 $\pm$ 0.715} & \underline{35.040 $\pm$ 0.768} \\
&                             & PolyFusion\_2M (Ours) & 0.894 $\pm$ 0.008 & 24.130 $\pm$ 1.330 & 36.020 $\pm$ 1.330 \\
&                             & PolyFusion\_5M (Ours) & \textbf{0.907 $\pm$ 0.004} & \textbf{22.450 $\pm$ 0.707} & \textbf{33.690 $\pm$ 0.640} \\
\hline

\multirow{8}{*}{$T_m$}  & \multirow{3}{*}{Unimodal}   & ChemBERTa \cite{chithrananda2020chemberta}   & 0.717 $\pm$ 0.033 & 39.840 $\pm$ 2.220 & 57.410 $\pm$ 3.240 \\
&                             & PolyCL \cite{zhou2025polycl}      & 0.581 $\pm$ 0.023 & 51.650 $\pm$ 1.320 & 70.000 $\pm$ 1.690 \\
&                             & PolyBERT \cite{kuenneth2023polybert}    & 0.711 $\pm$ 0.033 & 40.390 $\pm$ 2.510 & 58.060 $\pm$ 2.860 \\
\hhline{|~|-|-|-|-|-|}
& \multirow{5}{*}{Multimodal} & MMPolymer \cite{wang2024mmpolymer}   & 0.616 $\pm$ 0.020 & 50.150 $\pm$ 2.650 & 67.020 $\pm$ 2.100 \\
&                             & MoleculeSTM \cite{liu2023multimodal} & 0.556 $\pm$ 0.025 & 54.220 $\pm$ 0.648 & 72.020 $\pm$ 0.494 \\
&                             & PolyNC \cite{qiu2024polync}      & 0.742 $\pm$ 0.033 & 37.900 $\pm$ 2.140 & 54.780 $\pm$ 3.310 \\
&                             & PolyFusion\_2M (Ours) & \underline{0.747 $\pm$ 0.032} & \underline{37.300 $\pm$ 2.610} & \underline{54.300 $\pm$ 3.180} \\
&                             & PolyFusion\_5M (Ours) & \textbf{0.770 $\pm$ 0.021} & \textbf{34.800 $\pm$ 1.360} & \textbf{51.810 $\pm$ 1.810} \\
\hline

\multirow{8}{*}{$T_d$}  & \multirow{3}{*}{Unimodal}   & ChemBERTa \cite{chithrananda2020chemberta}   & 0.627 $\pm$ 0.022 & 45.890 $\pm$ 1.700 & 68.870 $\pm$ 2.790 \\
&                             & PolyCL \cite{zhou2025polycl}      & 0.528 $\pm$ 0.035 & 54.970 $\pm$ 2.750 & 77.420 $\pm$ 3.350 \\
&                             & PolyBERT \cite{kuenneth2023polybert}    & 0.648 $\pm$ 0.038 & 46.030 $\pm$ 3.470 & 66.800 $\pm$ 4.560 \\
\hhline{|~|-|-|-|-|-|}
& \multirow{5}{*}{Multimodal} & MMPolymer \cite{wang2024mmpolymer}   & 0.531 $\pm$ 0.038 & 55.650 $\pm$ 3.460 & 77.200 $\pm$ 3.960 \\
&                             & MoleculeSTM \cite{liu2023multimodal} & 0.482 $\pm$ 0.009 & 59.520 $\pm$ 1.420 & 81.130 $\pm$ 1.190 \\
&                             & PolyNC \cite{qiu2024polync}      & \underline{0.705 $\pm$ 0.017} & \underline{41.120 $\pm$ 1.420} & \underline{61.270 $\pm$ 2.360} \\
&                             & PolyFusion\_2M (Ours) & 0.682 $\pm$ 0.024 & 41.870 $\pm$ 1.870 & 63.540 $\pm$ 3.460 \\
&                             & PolyFusion\_5M (Ours) & \textbf{0.717 $\pm$ 0.027} & \textbf{39.570 $\pm$ 1.890} & \textbf{59.980 $\pm$ 3.750} \\
\hline

\end{tabular}%
}
\end{table*}

Importantly, these gains emerge before reaching the largest scale regime. PolyFusion\_2M (PolyFusion pretrained on 2M polymers) already tracks the strongest baselines across properties and narrows the performance gap relative to PolyFusion\_5M, indicating that alignment improves data efficiency: the representation generalizes by internalizing cross-view invariances rather than over-specializing to a single modality’s inductive bias. Overall, the transfer results support the design hypothesis of PolyFusion: a fused, contrastively aligned embedding space provides a more stable substrate for thermophysical prediction than any unimodal descriptor alone, and it benefits further—albeit with diminishing returns—from increased pretraining scale.

\subsection{Property-conditioned generation produces valid, novel and diverse polymers}
Forward prediction closes only the analysis loop; a more helpful FM must also close the design loop by generating candidates that (i) decode to chemically valid polymers, (ii) avoid degenerating into repeated templates, and (iii) explore genuinely new regions of chemistry space under an explicit property constraint. Using PolyFusion embeddings as the conditioning interface and the pretrained SELF-referencing Embedded Strings--Transformer Encoder--Decoder (SELFIES-TED) \cite{priyadarsini2024selfies} generative module, we assessed inverse design for $\rho$, $T_g$, $T_m$, and $T_d$ (Section~\ref{sec:generation}).

\begin{figure*}[htbp]
    \centering

    \begin{subfigure}{\textwidth}
        \centering
        \resizebox{\textwidth}{!}{%
        \begin{tabular}{|l|l|l|c|c|c|}
        \hline
        \textbf{Property} & \textbf{Input Setting} & \textbf{Model} &
        \textbf{Validity} (\%) $\uparrow$ & \textbf{Novelty} (\%) $\uparrow$ & \textbf{Diversity} $\uparrow$ \\
        \hline

        \multirow{8}{*}{$\rho$} 
        & \multirow{3}{*}{Unimodal} & ChemBERTa \cite{chithrananda2020chemberta}   & 99.934 $\pm$ 0.072 & 80.668 $\pm$ 4.624 & 0.829 $\pm$ 0.019 \\
        &                           & PolyCL \cite{zhou2025polycl}      & 99.995 $\pm$ 0.006 & \textbf{85.609 $\pm$ 2.902} & 0.828 $\pm$ 0.021 \\
        &                           & PolyBERT \cite{kuenneth2023polybert}    & 99.702 $\pm$ 0.518 & \underline{84.218 $\pm$ 2.410} & 0.819 $\pm$ 0.023 \\
        \hhline{|~|-|-|-|-|-|}
        & \multirow{5}{*}{Multimodal} & MMPolymer \cite{wang2024mmpolymer}   & \underline{99.998 $\pm$ 0.005} & 58.498 $\pm$ 7.352 & 0.523 $\pm$ 0.055 \\
        &                             & MoleculeSTM \cite{liu2023multimodal} & 99.985 $\pm$ 0.029 & 83.506 $\pm$ 3.362 & 0.816 $\pm$ 0.030 \\
        &                             & PolyNC \cite{qiu2024polync}      & 99.849 $\pm$ 0.142 & 82.325 $\pm$ 3.454 & \textbf{0.832 $\pm$ 0.026} \\
        &                             & PolyFusion\_2M (Ours)   & \textbf{100.000 $\pm$ 0.000} & 78.041 $\pm$ 3.181 & \underline{0.832 $\pm$ 0.028} \\
        &                             & PolyFusion\_5M (Ours)   & 99.885 $\pm$ 0.141 & 74.009 $\pm$ 1.984 & 0.828 $\pm$ 0.013 \\
        \hline

        \multirow{8}{*}{$T_g$} 
        & \multirow{3}{*}{Unimodal} & ChemBERTa \cite{chithrananda2020chemberta}   & 99.341 $\pm$ 0.216 & 85.405 $\pm$ 3.770 & 0.720 $\pm$ 0.060 \\
        &                           & PolyCL \cite{zhou2025polycl}      & 99.873 $\pm$ 0.119 & 86.779 $\pm$ 2.309 & 0.740 $\pm$ 0.021 \\
        &                           & PolyBERT \cite{kuenneth2023polybert}    & 99.048 $\pm$ 0.412 & 82.744 $\pm$ 4.884 & 0.714 $\pm$ 0.113 \\
        \hhline{|~|-|-|-|-|-|}
        & \multirow{5}{*}{Multimodal} & MMPolymer \cite{wang2024mmpolymer}   & \underline{99.890 $\pm$ 0.083} & 30.808 $\pm$ 4.471 & 0.284 $\pm$ 0.041 \\
        &                             & MoleculeSTM \cite{liu2023multimodal} & \textbf{99.951 $\pm$ 0.035} & 87.446 $\pm$ 6.227 & 0.735 $\pm$ 0.091 \\
        &                             & PolyNC \cite{qiu2024polync}      & 99.746 $\pm$ 0.310 & \underline{88.715 $\pm$ 9.670} & \underline{0.754 $\pm$ 0.100} \\
        &                             & PolyFusion\_2M (Ours)   & 99.785 $\pm$ 0.219 & 73.197 $\pm$ 17.447 & 0.592 $\pm$ 0.182 \\
        &                             & PolyFusion\_5M (Ours)   & 98.721 $\pm$ 0.710 & \textbf{91.251 $\pm$ 1.725} & \textbf{0.802 $\pm$ 0.015} \\
        \hline

        \multirow{8}{*}{$T_m$} 
        & \multirow{3}{*}{Unimodal} & ChemBERTa \cite{chithrananda2020chemberta}   & 99.729 $\pm$ 0.253 & 78.319 $\pm$ 6.699 & 0.689 $\pm$ 0.105 \\
        &                           & PolyCL \cite{zhou2025polycl}      & \underline{99.995 $\pm$ 0.010} & \underline{87.440 $\pm$ 3.391} & 0.754 $\pm$ 0.036 \\
        &                           & PolyBERT \cite{kuenneth2023polybert}    & 99.414 $\pm$ 0.299 & 82.125 $\pm$ 6.854 & \textbf{0.791 $\pm$ 0.034} \\
        \hhline{|~|-|-|-|-|-|}
        & \multirow{5}{*}{Multimodal} & MMPolymer \cite{wang2024mmpolymer}   & 99.797 $\pm$ 0.347 & 28.961 $\pm$ 13.932 & 0.277 $\pm$ 0.151 \\
        &                             & MoleculeSTM \cite{liu2023multimodal} & \textbf{99.998 $\pm$ 0.005} & 82.935 $\pm$ 4.243 & 0.772 $\pm$ 0.032 \\
        &                             & PolyNC \cite{qiu2024polync}      & 99.846 $\pm$ 0.190 & 84.744 $\pm$ 4.077 & 0.764 $\pm$ 0.025 \\
        &                             & PolyFusion\_2M (Ours)   & 99.985 $\pm$ 0.024 & 77.153 $\pm$ 5.973 & 0.683 $\pm$ 0.120 \\
        &                             & PolyFusion\_5M (Ours)   & 99.500 $\pm$ 0.497 & \textbf{89.477 $\pm$ 3.842} & \underline{0.785 $\pm$ 0.024} \\
        \hline

        \multirow{8}{*}{$T_d$} 
        & \multirow{3}{*}{Unimodal} & ChemBERTa \cite{chithrananda2020chemberta}   & 99.153 $\pm$ 0.470 & \underline{84.277 $\pm$ 3.388} & 0.715 $\pm$ 0.015 \\
        &                           & PolyCL \cite{zhou2025polycl}      & \textbf{99.985 $\pm$ 0.024} & 82.693 $\pm$ 6.509 & \underline{0.736 $\pm$ 0.042} \\
        &                           & PolyBERT \cite{kuenneth2023polybert}    & 98.972 $\pm$ 0.324 & 81.698 $\pm$ 3.940 & 0.726 $\pm$ 0.046 \\
        \hhline{|~|-|-|-|-|-|}
        & \multirow{5}{*}{Multimodal} & MMPolymer \cite{wang2024mmpolymer}   & \underline{99.966 $\pm$ 0.033} & 34.116 $\pm$ 2.605 & 0.319 $\pm$ 0.010 \\
        &                             & MoleculeSTM \cite{liu2023multimodal} & \underline{99.966 $\pm$ 0.036} & 77.166 $\pm$ 3.433 & 0.729 $\pm$ 0.063 \\
        &                             & PolyNC \cite{qiu2024polync}      & 99.182 $\pm$ 0.255 & 79.265 $\pm$ 3.817 & 0.645 $\pm$ 0.027 \\
        &                             & PolyFusion\_2M (Ours)   & 99.944 $\pm$ 0.053 & 69.979 $\pm$ 12.047 & 0.652 $\pm$ 0.206 \\
        &                             & PolyFusion\_5M (Ours)   & 99.299 $\pm$ 0.618 & \textbf{84.734 $\pm$ 3.555} & \textbf{0.792 $\pm$ 0.035} \\
        \hline
        \end{tabular}}
        \subcaption{}
        \label{fig:inverse_design_quality_and_novelty:A}
    \end{subfigure}

    \vspace{0.8em}

    \begin{subfigure}{\textwidth}
        \centering
        \includegraphics[width=\textwidth]{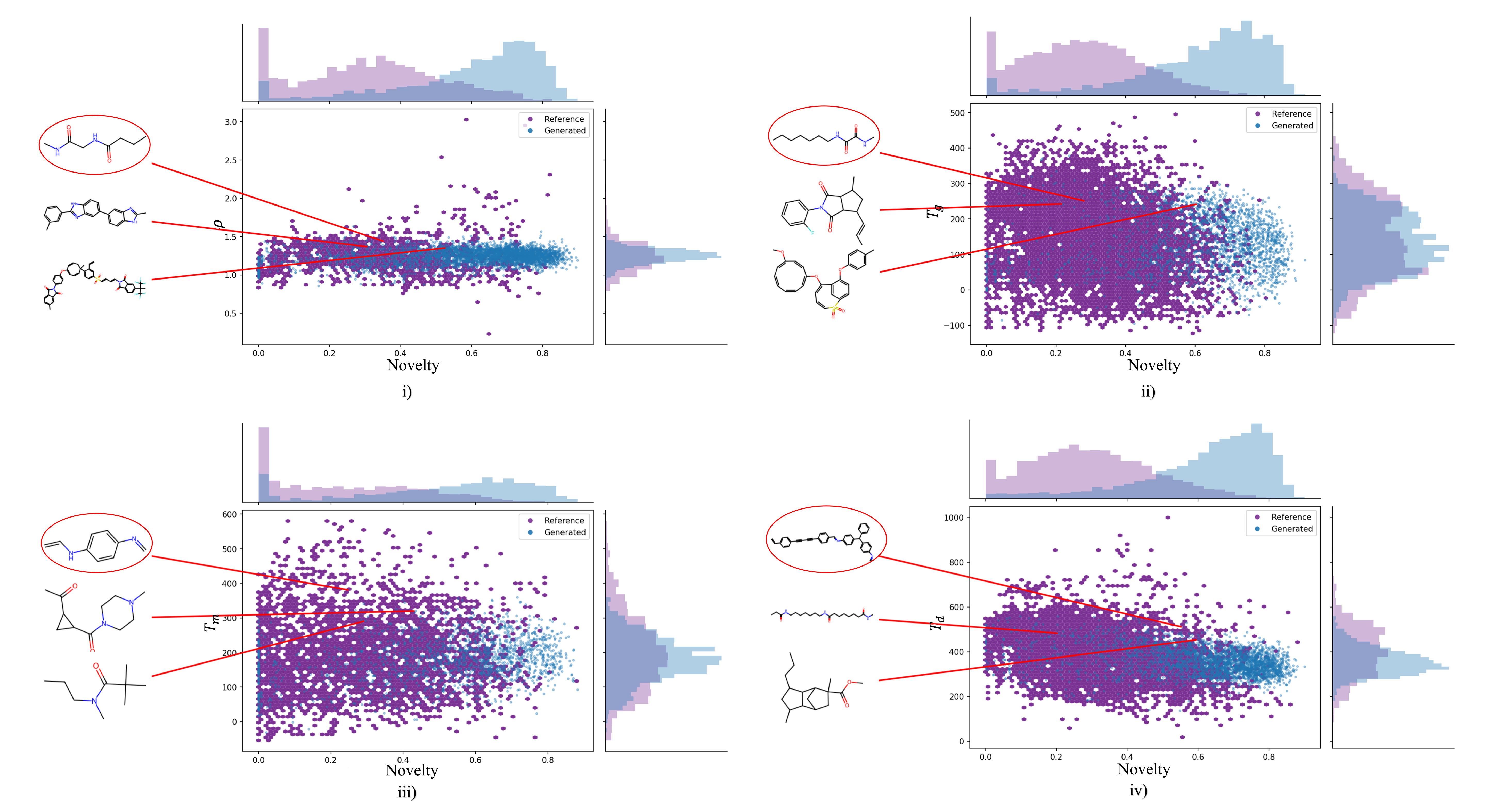}
        \subcaption{}
        \label{fig:inverse_design_quality_and_novelty:B}
    \end{subfigure}

    \caption{\textbf{Property-conditioned generation quality and novelty landscapes.}
    \textbf{A. Quantitative inverse-design performance.} Inverse-design performance for density (\(\rho\)), glass-transition temperature (\(T_g\)), melting temperature (\(T_m\)) and thermal-decomposition temperature (\(T_d\)) across unimodal and multimodal baselines and PolyFusion at two pretraining scales. For each property we report (i) validity, (ii) novelty, and (iii) diversity (Section~\ref{sec:generation}), with values given as mean \(\pm\) standard deviation over five cross-validation folds and the best/second-best for each metric highlighted in \textbf{bold}/\underline{underline}, respectively.
    \textbf{B. Novelty--property landscapes for PolyFusion\_5M.} For each property---(i) \(\rho\), (ii) \(T_g\), (iii) \(T_m\), and (iv) \(T_d\)---we compare the reference polymer set (purple; hexbin density) with PolyFusion\_5M generations (blue; scatter) by plotting novelty to reference (x-axis) versus the evaluated property value (y-axis), where novelty is the mean \(k\)-nearest-neighbor (kNN) Tanimoto distance \((1-\mathrm{sim})\) computed from Morgan fingerprints on PSMILES using leave-one-out reference-to-reference distances for the reference set and reference-set kNN distances for generated polymers. Marginal histograms (top/right) summarize novelty and property distributions for the reference and generated sets, and left callouts show representative structures (red-circled: reference; two accompanying: PolyFusion\_5M-generated).}
    \label{fig:inverse_design_quality_and_novelty}
\end{figure*}

Across methods and conditioned properties, generation validity is near-perfect (Fig.~\ref{fig:inverse_design_quality_and_novelty:A}), indicating that all models are capable of decoding latent representations into syntactically valid PSMILES. However, validity alone does not reflect generative quality. Several baseline models achieve high validity by producing structures that are overly conservative, leading to a pronounced reduction in novelty and diversity. In contrast, PolyFusion maintains near-perfect validity while simultaneously preserving substantially higher novelty and structural diversity, demonstrating that it better balances decodability with meaningful exploration of the polymer design space, especially conditioned on thermal transition and stability properties (\(T_g\), \(T_m\), and \(T_d\)).

An interesting observation is that the unimodal baseline PolyCL \cite{zhou2025polycl} generates the most novel polymers in the density (\(\rho\))-conditioned generation task (Fig.~\ref{fig:inverse_design_quality_and_novelty:A}), while maintaining high validity and diversity. However, this advantage does not extend to other conditioned properties such $T_g$, $T_m$, and $T_d$. This suggests that PolyCL’s less constrained latent space particularly benefits the generation of polymers conditioned on global, composition-driven properties like density, but is less effective for thermophysical properties that depend more strongly on fine-grained structural, geometric, and cross-modal information. In these cases, the additional constraints imposed by multimodal fusion appear to better regularize the generation process, leading to improved controllability and generalization.

To move beyond summary performance metrics and better understand where PolyFusion explores polymer design space under property constraints, we also examined the relationship between structural novelty and target properties for candidates generated by PolyFusion\_5M (Fig.~\ref{fig:inverse_design_quality_and_novelty:B}). Across all four properties, the generated polymers are systematically more dissimilar from the training set (larger nolvelty), while still remaining within physically reasonable ranges of the target properties. This indicates that PolyFusion explores new chemical structures without drifting into unrealistic or unphysical property regimes. Importantly, property conditioning narrows the spread of the target property, right histograms) while preserving high structural novelty.
For density (Fig.~\ref{fig:inverse_design_quality_and_novelty:B}(i)), the generated polymers cluster tightly around the desired $\rho$ value regardless of novelty, reflecting the fact that similar packing efficiency and cohesive energy density can be achieved through many different chemical architectures.
In contrast, for thermal properties (Fig.~\ref{fig:inverse_design_quality_and_novelty:B}(ii-iv)), PolyFusion preferentially samples polymers that are both highly novel and located near the upper end of $T_g$, $T_m$, or $T_d$ observed in the reference set. These regions correspond to sparsely populated areas of the training data, where desirable thermal performance exists but few examples are available. The representative repeat units highlighted in the figure combine rigid cyclic or aromatic segments with heteroatom-rich linkages and adjustable flexible spacers.
These motifs are consistent with established polymer design principles for increasing segmental stiffness and thermal stability, providing a physically interpretable explanation for how PolyFusion balances novelty and performance. Such behavior is critical for inverse polymer design: a useful generative model must not only produce chemically valid polymers, but also explore genuinely new regions of chemical space while remaining anchored to realistic, high-performance property regimes.

\subsection{PolyAgent couples PolyFusion with retrieval to produce grounded, constraint-consistent design actions} 
\begin{figure}[htbp]
    \centering
    \includegraphics[width=\textwidth]{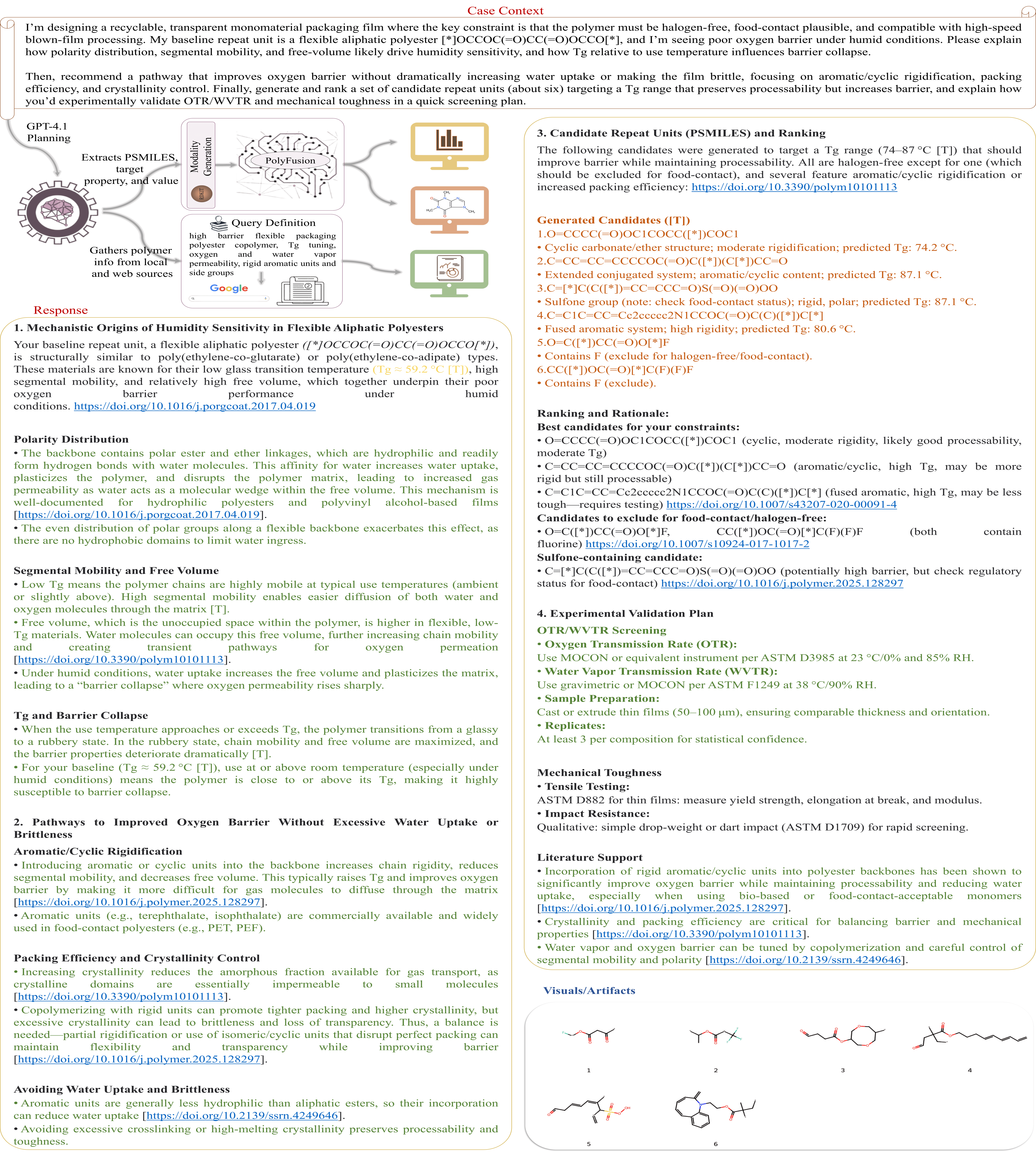}
    \caption{\textbf{PolyAgent: a tool-mediated workflow that converts open-ended design prompts into grounded, constraint-consistent actions.}
PolyAgent integrates an LLM controller with four domain tools to produce audit-ready polymer design outputs. \textbf{A.} Problem formulation and constraint extraction: free-form user requirements are parsed into typed constraints and decision variables that determine which tools are invoked and what constitutes an admissible candidate. \textbf{B.} Grounded mechanistic diagnosis: retrieval augments the response with evidence-linked support for mechanistic claims and constraint interpretations (blue: inline citations/links), while PolyFusion-based property prediction provides quantitative context for key operating regimes (yellow: predicted property callouts). \textbf{C.} Proposal, ranking, and verification: PolyFusion-based conditional generation proposes candidates consistent with the extracted constraints (orange: generated candidates and ranking rationale); follow-on prediction scores and ranks candidates within the same representation; and a structured validation protocol (e.g., barrier/mechanics screening under relevant conditions) is produced (green: verification plan), alongside molecular renderings for rapid expert audit and experimental follow-up.} 
    \label{fig:polyagent}
\end{figure}

Building on the transferable fused representations and property-conditioned generation established above, we introduced PolyAgent as a system-level interface that turns open-ended polymer design prompts into grounded and constraint-consistent recommendations by coupling PolyFusion with tool-mediated verification (Fig.\ref{fig:polyagent}). PolyAgent is implemented as a GPT-4.1 controller that decomposes a user request into a small set of typed sub-tasks and routes each sub-task to a dedicated tool: (i) a PolyFusion-based property predictor, (ii) a PolyFusion-based property-conditioned generator, (iii) a retrieval module that grounds mechanistic claims and regulatory/process constraints in external documents and databases via retrieval-augmented generation (RAG) and web search \cite{lewis2020retrieval}, and (iv) a structure visualizer that renders candidates into human-auditable artifacts. This design makes the large language model (LLM) decision-making layer explicit: rather than relying on purely parametric recall, PolyAgent treats evidence, candidate proposals, and verification as first-class outputs, and composes them into a report that separates explanation (mechanistic diagnosis), proposal (candidate chemistries and design levers), and verification (tests and acceptance criteria).

Across a suite of 20 simulated design cases spanning diverse objectives and constraints, PolyAgent produced higher-utility outputs by repeatedly enforcing three principles that are difficult to satisfy with an LLM alone. First, constraint operationalization: natural-language requirements (e.g., “food-contact plausible”, “halogen-free”, “monomaterial”, “process-compatible”) are converted into admissible chemical operations and explicit exclusion rules, which are carried forward to generation and ranking. Second, evidence traceability: mechanistic statements and constraint interpretations are tied to retrieved sources and surfaced as localized citations, enabling users to audit why specific design moves are recommended and under which assumptions they remain valid. Third, representation-consistent proposal and scoring: candidate polymers are generated and evaluated within the same shared multimodal embedding space provided by PolyFusion, avoiding the common failure mode where generation and prediction operate on incompatible featurizations and produce internally inconsistent rankings. In aggregate, this tool-mediated loop transforms PolyFusion from an offline representation model into a reproducible decision workflow: recommendations are not only plausible, but are also (i) grounded in accessible evidence, (ii) quantitatively prioritized by model-based scoring, and (iii) accompanied by an experiment-ready validation plan that closes the design loop.

\subsection{Tool access and retrieval improve reliability of the PolyAgent}
To isolate the contribution of tool access and retrieval to interactive polymer-design reliability, we benchmarked PolyAgent against two strong open-weight baselines (Mixtral-8$\times$22B \cite{jiang2024mixtral} and Llama-3.1-8B) operating without tool integration. Responses were scored along five practitioner-relevant axes—tool use, completeness, correctness, helpfulness, and citation accuracy—using a shared rubric (Section~\ref{sec:baselines}). For each query, five independent evaluators scored all systems, and we report mean scores aggregated over 20 simulated polymer-design queries (see Supplementary Note 7).

Figure~\ref{fig:polyagent_llm_benchmark}A illustrates a representative thermally focused query requesting an estimated decomposition onset temperature for a specified repeat unit. PolyAgent converts the request into a tool-backed prediction step, returns a concrete numerical estimate (e.g., a predicted $T_d$), evaluates it against the stated target, and then grounds mechanistic interpretation with localized, retrievable references. In contrast, the unaided baselines produce chemically plausible narrative commentary but exhibit two recurrent limitations. First, quantitative statements are under-constrained: outputs default to heuristic ranges, target-hugging point claims, or qualitative hedges without an explicit verification step. Second, evidential traceability is weak: citations are absent or not coupled to specific claims in a way that enables audit, leaving practitioners unable to determine whether mechanistic statements are supported or merely rhetorically plausible. PolyAgent addresses both issues by enforcing a tool-mediated verification loop that (i) invokes property prediction when a quantitative estimate is requested and (ii) uses retrieval to attach traceable sources to mechanistic and constraint-interpretation statements. This shifts the interaction from assumption-led “explain-and-suggest” behaviour toward “estimate-and-check” outputs in which numerical claims are explicitly tied to tool outputs and supporting statements are linked to verifiable evidence.

These behavioural differences are reflected in the aggregate scoring (Fig.~\ref{fig:polyagent_llm_benchmark}B; see Supplementary Note 8). PolyAgent shows a high, balanced profile across all five metrics, with near-ceiling Tool Use and consistently strong Citation Accuracy. In contrast, the tool-free baselines exhibit suppressed Tool Use and lower Completeness on queries where a numerical estimate is central, because they cannot execute an external prediction step and therefore often substitute invented quantitative content—heuristic ranges, target-anchored point values, or “order-of-magnitude” estimates. Consistent with this limitation, both baselines also remain low on Citation Accuracy: even when citations are provided, they are rarely coupled to specific claims in a way that permits verification of the quantitative assertions. Overall, Fig.\ref{fig:polyagent_llm_benchmark} supports a systems-level conclusion: in polymer design dialogue, reliability is driven less by narrative fluency than by whether the system can (i) invoke appropriate external procedures when quantitative claims are requested and (ii) attach traceable evidence to mechanistic and constraint-interpretation statements. By enforcing an explicit decomposition of the interaction into predict $\rightarrow$ evaluate against target $\rightarrow$ justify with retrieved evidence, PolyAgent produces responses that are more auditable and decision-relevant under practical design constraints.

\begin{figure}[htbp]
    \centering
    \includegraphics[width=\textwidth]{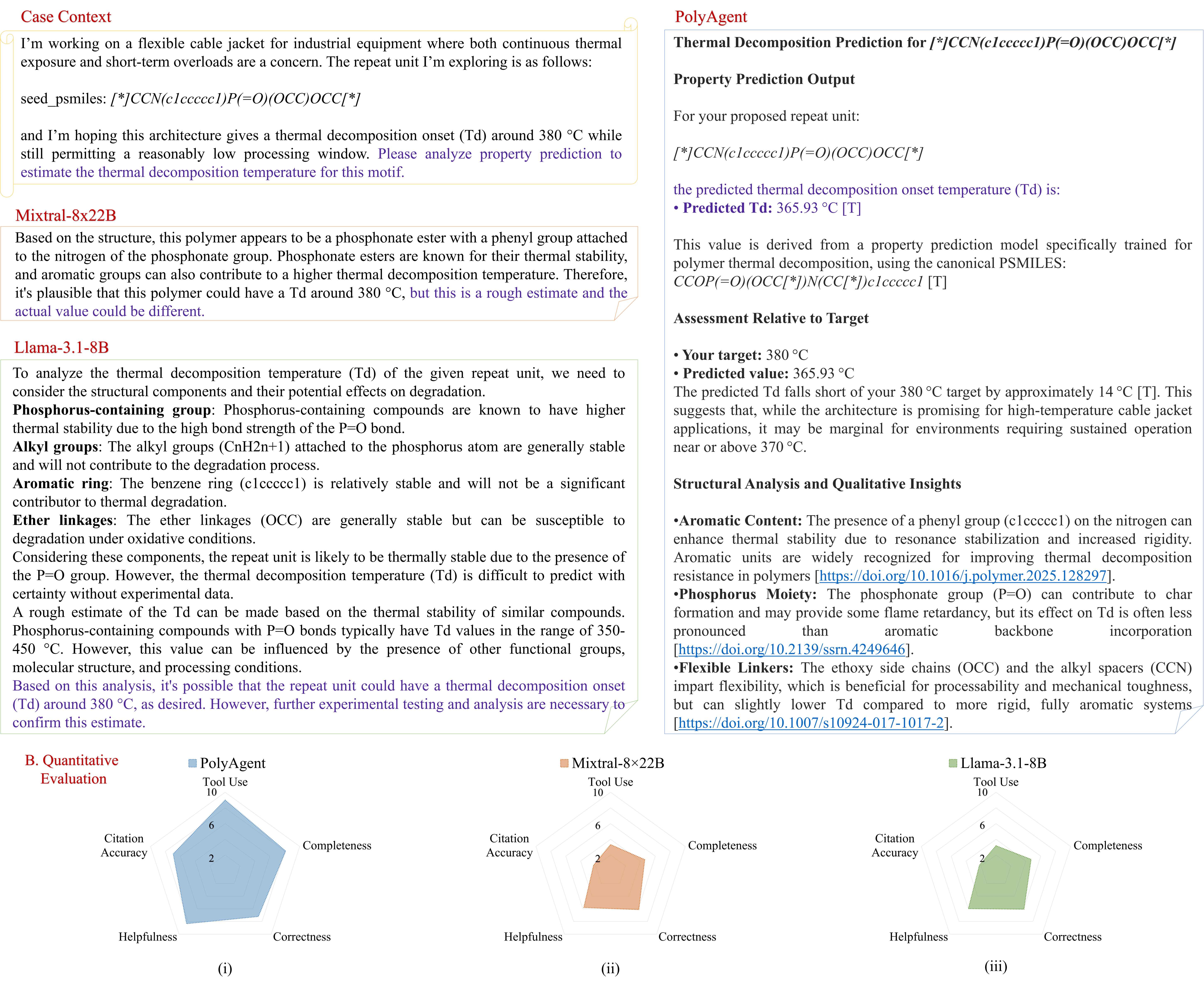}
\caption{\textbf{Tool access and retrieval improve the reliability of PolyAgent relative to open-weight LLM baselines.}
\textbf{A.} Representative comparison on a thermally focused design query. PolyAgent triggers a decomposition-prediction tool to produce a quantitative estimate, evaluates it against the stated target, and grounds mechanistic statements with localized, retrievable references. Mixtral-8$\times$22B and Llama-3.1-8B provide qualitatively plausible analyses but—without tool-backed verification—default to heuristic ranges or assumption-driven point claims with weaker evidence traceability. \textbf{B.} (i--iii) Radar plots of aggregate benchmark scores (0--10) across five metrics—Tool Use, Completeness, Correctness, Helpfulness, and Citation Accuracy—averaged over five independent evaluators and aggregated over $\sim$20 simulated cases (see Supplementary Note 7).}
    \label{fig:polyagent_llm_benchmark}
\end{figure}

\section{Discussion}
\label{sec:discussion}
This work positions polymer FMs as decision substrates rather than isolated predictors. PolyFusionAgent unifies multimodal pretraining (PolyFusion) with property prediction, property-conditioned generation, and a retrieval- and tool-augmented controller (PolyAgent) that produces audit-ready recommendations. The key advance is not only improved performance on downstream benchmarks, but a systems-level re-framing of polymer AI: design dialogue becomes reliable when quantitative claims are produced by explicit procedures (predictors/screeners) and mechanistic rationales are tethered to traceable sources, rather than inferred implicitly from a single representation or expressed as narrative plausibility. In this sense, PolyFusionAgent addresses a practical gap between polymer AI benchmarks and laboratory decision-making, where actions must be justified, constrained, and revisable.

Within this decision-oriented framing, the first requirement is a representation that remains stable under heterogeneous polymer inputs and dataset shift while retaining the information needed for both prediction and design. A central finding is that contrastively aligned multimodal embeddings transfer consistently to thermophysical prediction and remain navigable for conditional generation. This behaviour is informative from a polymer-physics perspective because the determinants of properties such as $T_g$, $T_m$, and $T_d$ are intrinsically multi-factorial and expressed at multiple scales. Local chemistry (functional groups, bond strengths, polarity and rigidity) sets interaction potentials and weak-link stability; packing and cohesive energy density shape free volume and transition behaviour; and conformational degrees of freedom mediate the accessibility of these interactions under thermal excitation. Any single encoding imposes an inductive bias that selectively amplifies part of this hierarchy. Sequence views are expressive but entangle syntax with chemistry; graphs preserve local neighbourhood structure but underrepresent longer-range context; approximate 3D proxies introduce useful distance-conditioned priors yet imperfectly reflect ensembles; fingerprints provide a stable similarity geometry but inherit truncation and hashing biases. PolyFusion leverages these complementary invariances by enforcing cross-view consistency during pretraining, which empirically reduces modality-specific distortion (Fig.\ref{fig:scale_fusion}) and yields a shared latent space that is less sensitive to missing or noisy information in any single view. Conceptually, alignment behaves as a form of multi-view regularization: the model is encouraged to encode property-relevant signals that are consistent across representations, thereby improving stability under heterogeneous polymer inputs and dataset shift.

A consequence of this stability is that transfer improvements become most apparent exactly where polymer physics is most coupled. The largest gains emerge for $T_g$, $T_m$, and $T_d$, properties that depend on coupled chemical--structural controls rather than on a single dominant descriptor. This pattern is consistent with the idea that multimodal fusion is most beneficial when prediction requires simultaneous sensitivity to (i) local motifs (e.g., labile bonds, aromaticity, heteroatom environments) and (ii) higher-order proxies for rigidity, packing frustration, and intermolecular constraint. In such regimes, unimodal representations are prone to underspecification: different polymers can appear similar under one view while differing meaningfully under another. A fused embedding can resolve these degeneracies by embedding polymers in a coordinate system shaped by multiple, partially independent constraints. The observation that a smaller-scale regime (2M) remains competitive further suggests that alignment can improve data efficiency by injecting redundant but complementary supervision, reducing reliance on any single corpus to provide complete coverage of relevant invariances.

Once a representation is transferable, the next requirement for a decision substrate is internal consistency across the tools that act on it. A persistent limitation in polymer design pipelines is that candidate generation, property scoring, and similarity/retrieval often operate over mismatched featurizations, producing rankings that are difficult to interpret or reproduce. PolyFusionAgent mitigates this by exposing a single embedding interface to prediction, generation, and corpus comparison. This coherence is operationally important: it allows the generator to be conditioned on the same representation that the predictor consumes, reducing the risk that candidates are ``good'' only under a generator-specific geometry. It also enables retrieval and nearest-neighbour contextualization to be performed in a consistent similarity space, which is essential for grounding decisions in precedent and for diagnosing when a candidate lies far from the training manifold where model uncertainty is expected to increase.

This representation-level coherence is also what makes inverse design behaviour meaningful beyond mere decodability. While validity saturates across methods, novelty and diversity distinguish whether a generator can explore property-consistent regions beyond common templates (Fig.\ref{fig:inverse_design_quality_and_novelty:B}). The favourable novelty--diversity profiles observed for PolyFusion, particularly for temperature-conditioned tasks, indicate that the fused latent space remains navigable under property constraints. Practically, this matters because inverse design is valuable only if conditioning does not collapse exploration to a narrow motif family. PolyFusion's ability to sustain multiple structural solutions to a single macroscopic target suggests that multimodal alignment improves the local geometry of the conditioning interface---making it easier to traverse the manifold without falling into degenerate modes---rather than merely improving decoding.

However, a coherent representation and a capable generator are still insufficient if the system cannot make its outputs verifiable in the way practitioners require. PolyAgent therefore targets a distinct failure mode that is common in LLM-only scientific assistants: fluent explanations coupled to under-constrained numbers and weak evidence traceability. The benchmark over $\sim$20 design cases shows that reliability gains are driven by two concrete mechanisms (Fig.\ref{fig:polyagent_llm_benchmark}). First, tool calls replace assumption-led numeric estimates with explicit predictions and target comparisons, improving correctness and completeness whenever a quantitative requirement is present. Second, retrieval enables localized citation linking for mechanistic claims and constraint interpretations, improving citation accuracy and making the rationale auditable. The result is a shift from ``explain-and-suggest'' outputs toward ``estimate-and-check'' outputs that expose a verifiable chain from evidence to recommendation. Importantly, this is a systems property: the improved profile does not require a larger base LLM, but rather a controller that can decide when to invoke external procedures and how to surface their outputs as decision-relevant artifacts.

Viewed together, the representation layer (PolyFusion) and the orchestration layer (PolyAgent) approximate the structure of real polymer design workflows, which are inherently iterative and knowledge-driven. PolyFusionAgent aligns with how polymer design decisions are typically made: hypotheses are generated in the context of precedent, feasibility constraints are applied early to avoid dead-end synthesis routes, and quantitative screening is used iteratively to converge on acceptable trade-offs. By structuring outputs as (i) constraint operationalization, (ii) quantitative evaluation, and (iii) evidence-linked justification, PolyAgent provides a practical scaffold for integrating expert feedback and for documenting why a candidate was proposed and under what assumptions it remains valid. This is particularly relevant for domains where failure modes are costly and multi-objective (e.g., thermal stability vs processability vs flame retardancy), and where decisions must be defensible to multidisciplinary teams.

These capabilities also clarify where the next advances are likely to come from: improving the fidelity of what is represented, strengthening what is verified, and tightening how uncertainty is managed. Several limitations define the next technical frontier. First, repeat-unit representations and approximate conformers necessarily under-specify polymer microstructure and processing history; extending the input space to include tacticity, end-group chemistry, polydispersity, and process descriptors would make the representation closer to the variables that control performance in practice \cite{stuart2023sizing, lin2019bigsmiles}. Second, using fingerprints as an explicit alignment target provides a stable similarity geometry but also imports the biases of engineered descriptors; systematic ablations on target choice and hybrid objectives will be important to disentangle the contribution of alignment from the inductive bias of the target space \cite{gurnani2023polymer, zagidullin2021comparative}. Third, tool augmentation improves auditability but introduces dependency on tool quality; integrating uncertainty quantification and reliability-aware policies (e.g., ``retrieve more'' or ``defer to expert'' when a candidate is far from the training distribution) would strengthen safety and usefulness in open-ended settings \cite{tang2025assessing, mozannar2020consistent}. Finally, coupling the agent loop to physics-based simulation or high-throughput experimentation would enable closed-loop refinement grounded in measured feedback, moving from evidence retrieval to evidence creation \cite{knox2022autonomous}.

Overall, PolyFusionAgent demonstrates that polymer foundation models become more useful when they are (i) trained to align complementary representations into a shared, transferable latent space and (ii) embedded within a tool- and retrieval-mediated workflow that makes quantitative claims and mechanistic rationales auditable. By combining multimodal pretraining with representation-consistent prediction and generation, and by operationalizing design dialogue through verification and evidence tracing, the framework advances polymer informatics toward decision workflows that are both scalable and experimentally actionable.

\section*{CRediT authorship contribution statement}
\textbf{Manpreet Kaur:} Writing – original draft, Visualization, Validation, Software, Methodology, Formal analysis. \textbf{Xingying Zhang:} Writing – review \& editing, Methodology, Formal analysis. \textbf{Qian Liu:} Writing – review \& editing, Supervision, Resources, Methodology, Funding acquisition, Conceptualization.

\section*{Declaration of competing interest}
The authors declare that they have no known competing finan-
cial interests or personal relationships that could have appeared to
influence the work reported in this paper.

\section*{Acknowledgements}
The authors acknowledge financial support from the Natural Sciences and Engineering Research Council of Canada (NSERC) through a Discovery Grant (RGPIN-2024-05844) to Q.L.

\section*{Data availability}
Pretrained model weights and related artifacts (tokenizers, encoder checkpoints, downstream heads, and inverse-design components) are available at \url{https://huggingface.co/kaurm43/polyfusionagent-weights}.

\section*{Code availability}
Source code for \textit{PolyFusionAgent} (including \textit{PolyFusion} and \textit{PolyAgent}) is available at \url{https://github.com/mk1426/PolyFusionAgent}. An interactive demonstration of the \textit{PolyAgent} interface is hosted at \url{https://huggingface.co/spaces/kaurm43/PolyFusionAgent}.

\backmatter





\end{document}


\title[Article Title]{PolyFusionAgent: A Multimodal Foundation Model and Autonomous AI Assistant for Polymer Property Prediction and Inverse Design}








\begin{center}
{\Large Supplementary Materials for}\\[10pt]

{\Large \textbf{PolyFusionAgent: A Multimodal Foundation Model and Autonomous AI Assistant for Polymer Property Prediction and Inverse Design}}\\[12pt]

Manpreet Kaur\textsuperscript{a}, Xingying Zhang\textsuperscript{b}, Qian Liu\textsuperscript{a,*}\\[6pt]

\textsuperscript{a}Department of Applied Computer Science, The University of Winnipeg, 515 Portage Avenue, Winnipeg, MB R3B 2E9, Canada.\\
\textsuperscript{b}Department of Mechanical Engineering, University of Manitoba, Winnipeg, MB R3T 2N2, Canada.\\[6pt]

$^{*}$Corresponding author. E-mail address: qi.liu@uwinnipeg.ca (Q. Liu). 
\end{center}

\vspace{1em}

\noindent \textbf{This PDF file includes:}\\
Supplementary Text\\
Tables S1 to S3\\
References

\vspace{1.5em}
\newpage
\subsection*{Supplementary Note 1 - Foundation models for polymer informatics}
\label{supp:note1}
Polymer informatics is increasingly converging on the foundation-model (FM) paradigm \cite{ferji2025basic, takeda2023foundation}: pretraining high-capacity neural encoders on large, weakly labeled or self-supervised polymer corpora to learn transferable representations that support many downstream tasks with limited additional supervision. In polymers, the central FM challenge is not only scale, but representation mismatch---repeat units under-specify tacticity, processing history, dispersity, and morphology---so polymer FMs have tended to emphasize (i) robust string/graph formalisms (PSMILES/BigSMILES \cite{lin2019bigsmiles}/PSELFIES \cite{krenn2020selfies}), (ii) architectural inductive bias (Transformers for sequences; GNNs for connectivity; 3D networks for geometry) \cite{zhou2023unimol}, and (iii) alignment objectives that make embeddings comparable across modalities \cite{wang2022molecular, radford2021learning}. A practical consequence is that polymer FMs are increasingly designed as general-purpose encoders (for prediction/retrieval) \cite{kim2018polymer, ward2016general} and conditional generators (for inverse design) \cite{jin2018junction}, often coupled through a shared latent space for controllable search and optimization. In PolyFusion, this FM philosophy is explicitly operationalized by aligning four polymer views---PSMILES, 2D graphs, 3D conformers, and fingerprints---into a unit-normalized shared space, with standardized preprocessing to keep polymer repeat-unit chemistry compatible with conventional cheminformatics tooling.

\subsection*{Supplementary Note 2 - Foundation models for polymer property prediction}
\label{supp:note2}
Early polymer machine learning (ML) platforms focused on curated structure--property datasets and task-specific descriptors; the FM shift replaces hand-engineered featurization with learned polymer embeddings that can be adapted across many properties \cite{ramprasad2017machine, pilania2013accelerating}. Specifically, polyBERT \cite{kuenneth2023polybert} exemplifies this: a polymer language model pretrained at scale on PSMILES (utilizing the polyOne corpus \cite{kuenneth2022polyone}) and then fine-tuned for fast property inference, demonstrating that polymer-specific tokenization plus large-scale pretraining can match or surpass conventional fingerprints while dramatically accelerating inference throughput. A parallel direction is TransPolymer \cite{xu2023transpolymer}, which introduced chemically aware tokenization for polymer sequences and showed consistent gains across multiple property benchmarks, reinforcing the message that polymer sequence modeling benefits from polymer-aware segmentation rather than generic SMILES tokenization. Beyond pure ``chemical language,'' prompted or hybrid language formulations have emerged: PolyNC \cite{qiu2024polync} casts polymer property prediction as a text-to-text task combining natural-language prompts with chemical strings, enabling unified handling of regression and classification while improving robustness and multitask generalization.

In parallel, the Ramprasad group's broader ecosystem illustrates how FM-like scaling can occur outside Transformers. Polymer Genome \cite{kim2018polymer} established a data-powered property prediction capability and has continued to evolve as a platform that operationalizes rapid predictions over diverse polymer properties, providing an important baseline for what ``general-purpose'' polymer predictors must achieve in practice. At the graph-learning frontier, multitask graph neural network (GNN) scaling has been demonstrated for polymers, showing that directly learning from repeat-unit graphs can yield 1--2 orders-of-magnitude speedups in feature extraction while maintaining strong predictive fidelity across many properties. Most recently, polymer property prediction has expanded to multimodal pretraining as an explicit analogue of molecular multimodal FMs: frameworks such as MMPolymer \cite{wang2024mmpolymer} align 1D polymer strings with 3D structural information (including strategies to cope with the scarcity and ambiguity of polymer 3D), and Uni-Poly \cite{huang2025unified} integrates diverse polymer modalities into a unified representation that improves accuracy and applicability across tasks. Finally, polymer property prediction is increasingly influenced by general molecular FM advances (e.g., SMILES Transformers) as transferable initialization---useful when polymer data are limited---while polymer-specific curation and representation remain the dominant performance levers in-domain.

\subsection*{Supplementary Note 3 - Generative AI for inverse polymer design}
\label{supp:note3}
Inverse design in polymers operationalizes a different set of FM requirements: beyond learning structure--property correlations, the model must generate chemically valid, syntactically correct polymer representations while enabling property conditioning and optimization over vast, discontinuous search spaces. Several families of approaches now exist. (i) Sequence-based generative models: PolyTAO \cite{qiu2024ondemand} demonstrates on-demand generation conditioned on target properties at high validity, emphasizing pretraining plus property conditioning as an effective route to controllable generation. polyBART \cite{savit2025polybart} extends this concept with PSELFIES enabling bidirectional translation between structures and properties and positioning polymer generation as a language task that can be experimentally grounded. polyT5 \cite{sahu2026polyt5} (encoder--decoder) further generalizes conditioning and generation in a text-to-text framework that unifies understanding and controllable synthesis in one architecture. (ii) Latent-variable generative modeling: variational autoencoders (VAEs) have been used to explore complex polymer design subspaces, including topology-conditioned generation and property guidance, providing a principled way to trade off reconstruction fidelity and novelty in polymer chemical space. (iii) Graph-to-graph and structured generation: polyG2G \cite{gurnani2021polyg2g} is representative of generative design frameworks that directly learn mappings enabling polymer proposal generation for targeted applications such as dielectrics, highlighting the value of explicit structure manipulation rather than purely autoregressive string sampling.

A consistent theme across these approaches is that ``generation'' alone is insufficient for inverse design; practical pipelines increasingly incorporate optimization, validation, and benchmarking layers. Reinforcement learning (RL) is becoming a dominant mechanism for directional search and multi-objective trade-offs \cite{li2026polyrl}, including closed-loop variants that incorporate physical validation (e.g., gas separation polymer discovery). The community has also begun to standardize evaluation through benchmarking studies that compare VAE, graph adversarial network (GAN)-family, recurrent neural networks (RNN), and RL-augmented generators under common metrics (validity, novelty, diversity, property targeting error) \cite{yue2025benchmarking}, clarifying where model classes succeed or fail for polymer-specific constraints. In this landscape, PolyFusion's inverse-design approach is best positioned as a representation-first design strategy: by constructing a shared latent geometry aligned across modalities, downstream conditional generators and search methods can operate in a space where neighborhood structure is chemically meaningful, thereby reducing the brittleness that often arises when conditioning directly on sparse labels or on a single representation view.

\subsection*{Supplementary Note 4 - Dataset curation across PI1M, polyOne, and PolyInfo}
\label{supp:note4}
Polymer FMs are unusually sensitive to dataset provenance because the dominant sources of scale (hypothetical enumeration and property imputation) and the dominant sources of “truth” (experimentally reported measurements) encode different biases. A defensible curation strategy therefore separates representation learning at breadth from evaluation at fidelity: large unlabeled hypothetical corpora regularize tokenization/grammar, repeat-unit chemistry, and long-tail motif coverage, while smaller experimentally grounded corpora stress-test robustness to reporting heterogeneity, processing-condition confounding, and measurement noise.

This division maps directly onto how the three major resources were assembled. PI1M \cite{ma2020pi1m} was originally created to produce benchmark-scale PSMILES: a generative model was trained on roughly 12,000 PoLyInfo-curated polymers and then sampled to yield about 1M hypothetical polymers, intentionally densifying sparse regions of polymer chemical space and improving coverage of uncommon repeat units and backbone topologies. polyOne \cite{kuenneth2022polyone} instead targets the scaling regime for polymer language modeling — it supplies an enormous synthetic corpus ($\approx 100$ million hypothetical polymers) designed to expose models to a vast variety of token sequences and long-range patterns; because of its focus on scale, polyOne is especially useful when the training objective is to learn statistical regularities and emergent syntax at large data scale rather than to mirror experimental reporting. PoLyInfo \cite{otsuka2011polyinfo} is usually treated as the high-fidelity reference layer rather than a scaling substrate: it is literature-derived, condition-aware, and spans broad polymer classes (including blends and composites), which makes it well suited for downstream evaluation and domain adaptation. A statistical summary reported for January 2024 quantifies the database composition: 18,697 homopolymers, 7,737 copolymers, 2,572 polymer blends, and 3,069 composites, with 161,464 polymer samples underpinning the recorded measurements.

To train and probe PolyFusion at different pretraining scales while preserving the distinct roles of synthetic and experimental resources, we constructed two pretraining subsets drawn from the union of PI1M and polyOne: a medium-scale set containing 2M PSMILES and a large-scale set containing 5M PSMILES. Both subsets were sampled to preserve class diversity (homopolymer / copolymer / blend / composite proportions), repeat-unit frequency tails, and the token-length distribution so that common and rare chemistries are represented consistently at each scale. PolyInfo was held out from pretraining and reserved exclusively for fine-tuning and evaluation. Pretrained models were fine-tuned on measured property records drawn from PolyInfo for four thermophysical targets: density (\(\rho\)), glass-transition temperature (\(T_g\)), melting temperature (\(T_m\)), and thermal-decomposition temperature (\(T_d\)). The number of PoLyInfo samples used for each target was \(N_{\rho}=1756\), \(N_{T_g}=8191\), \(N_{T_m}=3902\), and \(N_{T_d}=6395\). 

\subsection*{Supplementary Note 5 - Unimodal Encoders in PolyFusion}
\label{supp:note5}

PolyFusion instantiates four unimodal encoders, one per modality \(\ast\in\{D,G,S,T\}\): \(D\) (PSMILES; DeBERTaV2), \(G\) (2D molecular graph; GINE), \(S\) (3D conformer; SchNet), and \(T\) (fingerprints; Transformer). For each polymer instance, encoder \(\ast\) produces a pooled hidden representation \(H_{\ast}\in\mathbb{R}^{d_h}\), which is then mapped to the shared space by a modality-specific projection head \(f_{\mathrm{proj}}(\cdot)\) to yield \(G_{\ast}\in\mathbb{R}^{d}\) (with \(d=600\)).

\begin{itemize}
  \item \textbf{DeBERTaV2 (\(D\)):}
  PSMILES sequences are encoded with a 12-layer DeBERTaV2-style Transformer (number of layers \(L=12\), model dimension \(d_{\text{model}}=600\), number of attention heads \(M=12\), per-head dimension \(d_{\text{head}}=d_{\text{model}}/M\), and intermediate (FFN) dimension 512) using disentangled attention with relative positional encodings. Let the tokenized PSMILES sequence be \(\mathbf{t}=(t_1,\ldots,t_T)\), where \(T\) is the number of tokens and \(t_i\) is the token index at position \(i\). Each token is mapped to an embedding
  \begin{equation}
  x_i = \mathbf{e}_{t_i} + \mathbf{p}_i,
  \qquad \mathbf{e}_{t_i}\in\mathbb{R}^{d_{\text{model}}},\ \mathbf{p}_i\in\mathbb{R}^{d_{\text{model}}}
  \label{eq:psmiles_embed}
  \end{equation}
  where \(\mathbf{e}_{t_i}\) is the learned token embedding and \(\mathbf{p}_i\) is the learned absolute positional embedding. Stacking token embeddings yields \(\mathbf{X}^{(0)}=[x_1;\ldots;x_T]\in\mathbb{R}^{T\times d_{\text{model}}}\), where \([\cdot;\cdot]\) denotes row-wise concatenation. For Transformer layer \(\ell\in\{0,\ldots,L-1\}\), \(\mathbf{X}^{(\ell)}\in\mathbb{R}^{T\times d_{\text{model}}}\) denotes the layer-\(\ell\) hidden states.

  For attention head \(m\in\{1,\ldots,M\}\), define content-based queries/keys/values
  \begin{equation}
  \mathbf{Q}^c = \mathbf{X}^{(\ell)}W_{Q,m}^c,\qquad
  \mathbf{K}^c = \mathbf{X}^{(\ell)}W_{K,m}^c,\qquad
  \mathbf{V}   = \mathbf{X}^{(\ell)}W_{V,m}
  \label{eq:deberta_qkv_content}
  \end{equation}
  where \(W_{Q,m}^c,W_{K,m}^c,W_{V,m}\in\mathbb{R}^{d_{\text{model}}\times d_{\text{head}}}\) are learned projection matrices, and \(\mathbf{Q}^c,\mathbf{K}^c,\mathbf{V}\in\mathbb{R}^{T\times d_{\text{head}}}\). Let \(\mathbf{Q}^{c,(m)}_i\in\mathbb{R}^{d_{\text{head}}}\) denote the query vector at position \(i\) for head \(m\), and similarly \(\mathbf{K}^{c,(m)}_j\in\mathbb{R}^{d_{\text{head}}}\) and \(\mathbf{V}^{(m)}_j\in\mathbb{R}^{d_{\text{head}}}\) denote the key and value at position \(j\) for head \(m\).

  Relative-position keys \(\mathbf{K}^r\) are obtained from a learned relative embedding table \(\mathbf{R}\in\mathbb{R}^{(2K+1)\times d_{\text{head}}}\), where \(K\in\mathbb{N}\) is the maximum relative distance. For query position \(i\) and key position \(j\),
  \begin{equation}
  \mathbf{K}^r_{i,j} = \mathbf{R}_{\mathrm{clip}(j-i,\,-K,\,K)}
  \label{eq:deberta_relpos}
  \end{equation}
  where \(\mathrm{clip}(u,-K,K)=\max(-K,\min(u,K))\). Let \(\mathbf{K}^{r,(m)}_{i,j}\in\mathbb{R}^{d_{\text{head}}}\) denote the head-\(m\) slice of \(\mathbf{K}^r_{i,j}\). The head-\(m\) attention logits for query position \(i\) and key position \(j\) are
  \begin{equation}
  a^{(m)}_{i,j}
  =
  \frac{1}{\sqrt{d_{\text{head}}}}
  \left(
  \langle \mathbf{Q}^{c,(m)}_i,\mathbf{K}^{c,(m)}_j\rangle
  +
  \langle \mathbf{Q}^{c,(m)}_i,\mathbf{K}^{r,(m)}_{i,j}\rangle
  \right)
  \label{eq:deberta_attn_logits}
  \end{equation}
  where \(\langle\cdot,\cdot\rangle\) denotes the Euclidean inner product. Attention weights are computed by normalizing over key positions,
  \(\alpha^{(m)}_{i,j}=\mathrm{softmax}_j(a^{(m)}_{i,j})=\exp(a^{(m)}_{i,j})/\sum_{j'=1}^{T}\exp(a^{(m)}_{i,j'})\),
  and head outputs are
  \begin{equation}
  \mathrm{head}^{(m)}_i = \sum_{j=1}^{T} \alpha^{(m)}_{i,j}\,\mathbf{V}^{(m)}_j.
  \label{eq:deberta_head_out}
  \end{equation}
  Concatenating the \(M\) heads and applying an output projection yields \(\mathrm{MHA}(\mathbf{X}^{(\ell)})\in\mathbb{R}^{T\times d_{\text{model}}}\). Each Transformer block uses residual connections and layer normalization:
  \begin{align}
  \tilde{\mathbf{X}}^{(\ell)} &= \mathrm{LN}\!\left(\mathbf{X}^{(\ell)} + \mathrm{MHA}\!\left(\mathbf{X}^{(\ell)}\right)\right),
  \label{eq:deberta_block_attn}\\
  \mathbf{X}^{(\ell+1)} &= \mathrm{LN}\!\left(\tilde{\mathbf{X}}^{(\ell)} + \mathrm{FFN}\!\left(\tilde{\mathbf{X}}^{(\ell)}\right)\right),
  \label{eq:deberta_block_ffn}
  \end{align}
  where \(\mathrm{LN}(\cdot)\) denotes layer normalization and \(\mathrm{FFN}(\cdot)\) is a position-wise feedforward network with intermediate dimension 512. Token-level hidden states from the final layer \(\mathbf{X}^{(L)}\) are mean-pooled to form the pooled representation
  \begin{equation}
  H_{D} = \frac{1}{T}\sum_{i=1}^{T} \mathbf{X}^{(L)}_i \in \mathbb{R}^{d_{\text{model}}},
  \label{eq:psmiles_pool_HD}
  \end{equation}
  where \(\mathbf{X}^{(L)}_i\in\mathbb{R}^{d_{\text{model}}}\) denotes the final-layer hidden state at token position \(i\). The pooled representation is projected to the shared space as \(G_{D}=f_{\mathrm{proj}}(H_{D})\in\mathbb{R}^{600}\), followed by \(\ell_2\)-normalization.

  \item \textbf{GINE (\(G\)):}
  For the GINE encoder, at message-passing layer \(\ell\in\{0,\ldots,L-1\}\), node states \(h_v^{(\ell)}\in\mathbb{R}^{d_g}\) are updated as
  \begin{equation}
  h_v^{(\ell+1)} =
  \mathrm{MLP}^{(\ell)}
  \left(
  h_v^{(\ell)} +
  \sum_{u\in\mathcal{N}(v)}
  \mathrm{ReLU}\!\left(h_u^{(\ell)} + \phi_e(e_{uv})\right)
  \right),
  \label{eq:gine_update}
  \end{equation}
  where \(v\) and \(u\) are nodes (atoms), \(\mathcal{N}(v)\) is the neighbor set of \(v\), \(e_{uv}\) denotes the edge (bond) features on \((u,v)\), \(\phi_e(\cdot)\) is a learnable edge embedding network mapping \(e_{uv}\) to \(\mathbb{R}^{d_g}\), \(\mathrm{ReLU}(\cdot)\) is the rectified linear unit, and \(\mathrm{MLP}^{(\ell)}(\cdot)\) is a learnable multilayer perceptron. We use \(L=5\) GINE layers with hidden size \(d_g=300\). Let \(\mathcal{V}\) denote the set of graph nodes and \(|\mathcal{V}|\) its cardinality. The pooled graph representation is obtained by mean-pooling node states from the final layer \(h_v^{(L)}\),
  \begin{equation}
  H_{G} = \frac{1}{|\mathcal{V}|}\sum_{v\in\mathcal{V}} h_v^{(L)} \in \mathbb{R}^{d_g},
  \label{eq:gine_pool_HG}
  \end{equation}
  and projected as \(G_{G}=f_{\mathrm{proj}}(H_{G})\in\mathbb{R}^{600}\), followed by \(\ell_2\)-normalization.

  \item \textbf{SchNet (\(S\)):}
  The 3D encoder is SchNet, which uses continuous-filter convolutions based on interatomic distances. Given atom features and positions \(\{(x_i,r_i)\}_{i=1}^{N}\), where \(N\) is the number of atoms, \(x_i\in\mathbb{R}^{d_s}\) is the input atom feature vector, and \(r_i\in\mathbb{R}^{3}\) is the Cartesian coordinate, SchNet constructs a neighbor list \(\mathcal{N}_r(v)\) within a radial cutoff \(r_{\mathrm{cut}}=10\,\text{\AA}\) with at most 64 neighbors per atom. At interaction layer \(\ell\in\{0,\ldots,L-1\}\), atom representations \(h_v^{(\ell)}\in\mathbb{R}^{d_s}\) are updated via
  \begin{equation}
  h_v^{(\ell+1)} = h_v^{(\ell)} + \sum_{u\in\mathcal{N}_r(v)}
  h_u^{(\ell)} \odot
  W_{\mathrm{filter}}^{(\ell)}\!\left(\psi(\|r_u-r_v\|_2)\right),
  \label{eq:schnet_update}
  \end{equation}
  where \(\odot\) denotes element-wise multiplication, \(\|r_u-r_v\|_2\) is the Euclidean distance between atoms \(u\) and \(v\), \(\psi(\cdot)\) is a Gaussian radial basis expansion of the distance, and \(W_{\mathrm{filter}}^{(\ell)}(\cdot)\) is a learnable filter network that maps the expanded distance features to \(\mathbb{R}^{d_s}\). We use \(L=6\) interaction layers with hidden size \(d_s=600\). Atom states are mean-pooled to obtain
  \begin{equation}
  H_{S} = \frac{1}{N}\sum_{v=1}^{N} h_v^{(L)} \in \mathbb{R}^{d_s},
  \label{eq:schnet_pool_HS}
  \end{equation}
  then projected as \(G_{S}=f_{\mathrm{proj}}(H_{S})\in\mathbb{R}^{600}\), followed by \(\ell_2\)-normalization.

  \item \textbf{Transformer (\(T\)):}
  Fingerprint vectors are represented as length-\(L=2048\) bit sequences \(\mathbf{b}=(b_1,\ldots,b_L)\), where \(b_i\in\{0,1\}\) is the bit value at position \(i\). Each position \(i\) is embedded using a learned bit-value embedding \(\mathbf{e}_{b_i}\) and a learned positional embedding \(\mathbf{p}_i\):
  \begin{equation}
  x_i = \mathbf{e}_{b_i} + \mathbf{p}_i,
  \qquad \mathbf{e}_{b_i},\mathbf{p}_i\in\mathbb{R}^{256}
  \label{eq:fp_embed}
  \end{equation}
  yielding \(\mathbf{X}^{(0)}=[x_1;\ldots;x_L]\in\mathbb{R}^{L\times 256}\). The sequence \((x_1,\ldots,x_L)\) is processed by a 4-layer Transformer encoder (number of layers \(L_T=4\)) with 8 attention heads and feedforward dimension 1024, producing final hidden states \(\{h_i\}_{i=1}^{L}\) with \(h_i\in\mathbb{R}^{256}\). These hidden states are mean-pooled to obtain
  \begin{equation}
  H_{T} = \frac{1}{L}\sum_{i=1}^{L} h_i \in \mathbb{R}^{256},
  \label{eq:fp_pool_HT}
  \end{equation}
  then projected via \(G_{T}=f_{\mathrm{proj}}(H_{T})\in\mathbb{R}^{600}\), followed by \(\ell_2\)-normalization.
\end{itemize}

\subsection*{Supplementary Note 6 - Ablation of unimodal encoders and pretraining scale}
\label{supp:note6}
\textbf{Unimodal property prediction (Table \ref{tab:regression_unimodal}):}
The unimodal ablation indicates clear, modality-specific strengths and heterogeneous sensitivity to pretraining scale. Sequence-native encoding of repeat-unit strings (DeBERTaV2 on PSMILES) consistently provides the most reliable single-view forward predictions across the evaluated thermophysical targets; its representations generalize robustly across targets and exhibit comparatively high predictive fidelity. Structure-native encoders (GINE, SchNet) are more variable in their response to increased pretraining: for some properties and metrics they benefit noticeably from larger pretraining corpora, while for others gains are marginal or absent. This heterogeneity suggests that graph- and geometry-based repeat-unit proxies encode useful structural information but are more dependent on scale to calibrate coverage of the long tail of chemotypes and to learn invariances to proxy noise (e.g., single-conformer approximations). The fingerprint-based Transformer (engineered descriptors) shows a different profile: its representations are comparatively stable across runs and scales yet exhibit a lower predictive ceiling than the best sequence model. Mechanistically, this pattern is consistent with fixed hashed descriptors imposing an information bottleneck that confers robustness but limits expressivity for targets where longer-range context or subtle topology–chemistry interactions matter. In sum, the regression results support the view that sequence encoders are the strongest unimodal predictors in this setting, structure-native encoders are more data-hungry and property-dependent, and descriptor-based encoders trade expressivity for stability.

\begin{table*}[h!]
\centering
\caption{\textbf{Effect of pretraining scale and modality-specific encoders on thermophysical property prediction.}
We report performance on four targets—density ($\rho$), glass-transition temperature ($T_g$), melting temperature ($T_m$), and thermal decomposition temperature ($T_d$)—using three metrics shown as mean $\pm$ standard deviation: $R^2$ (higher is better), mean absolute error (MAE; lower is better), and root-mean-square error (RMSE; lower is better). Results are given for models pretrained on two corpus scales (2 million, 2M; and 5 million, 5M polymers), with each block of rows in the table corresponding to one pretraining scale. The unimodal models compared are GINE (2D graph encoder), SchNet (3D conformation–aware encoder), DeBERTaV2 (PSMILES-based encoder), and a Transformer trained on engineered fingerprints. Values are the mean across five cross-validation runs and the reported uncertainty is the standard deviation across those runs. Boldface indicates the best-performing model for a given target and pretraining scale, and underlining indicates the second best. Overall, DeBERTaV2 consistently attains the strongest performance across targets and scales, and increasing pretraining scale from 2M to 5M generally improves or maintains performance, with the largest gains observed for structure-native encoders (GINE, SchNet).}
\label{tab:regression_unimodal}

\resizebox{\textwidth}{!}{%
\begin{tabular}{|l|l|l|c|c|c|}
\hline
\textbf{Property} & \textbf{Corpora Scale} & \textbf{Model} & \textbf{$R^2$ ($\uparrow$)} & \textbf{MAE ($\downarrow$)} & \textbf{RMSE ($\downarrow$)} \\
\hline

\multirow{8}{*}{$\rho$} 
& \multirow{4}{*}{2M} & GINE        & $0.705 \pm 0.078$ & $0.064 \pm 0.008$ & $0.110 \pm 0.018$ \\
&                     & SchNet      & $0.718 \pm 0.040$ & $0.062 \pm 0.003$ & $0.108 \pm 0.012$ \\
&                     & DeBERTaV2   & \underline{$0.742 \pm 0.060$} & \underline{$0.056 \pm 0.004$} & \underline{$0.103 \pm 0.016$} \\
&                     & Transformer & $0.526 \pm 0.030$ & $0.080 \pm 0.006$ & $0.140 \pm 0.009$ \\
\hhline{|~|-|-|-|-|-|}
& \multirow{4}{*}{5M} & GINE        & $0.700 \pm 0.046$ & $0.060 \pm 0.004$ & $0.111 \pm 0.013$ \\
&                     & SchNet      & $0.720 \pm 0.039$ & $0.061 \pm 0.003$ & $0.107 \pm 0.012$ \\
&                     & DeBERTaV2   & \bm{$0.744 \pm 0.058$} & \bm{$0.055 \pm 0.004$} & \bm{$0.102 \pm 0.015$} \\
&                     & Transformer & $0.543 \pm 0.033$ & $0.076 \pm 0.006$ & $0.138 \pm 0.010$ \\
\hline

\multirow{8}{*}{$T_g$} 
& \multirow{4}{*}{2M} & GINE        & $0.858 \pm 0.016$ & $30.065 \pm 2.067$ & $41.690 \pm 2.224$ \\
&                     & SchNet      & $0.860 \pm 0.007$ & $29.074 \pm 0.894$ & $41.438 \pm 0.968$ \\
&                     & DeBERTaV2   & \underline{$0.902 \pm 0.007$} & \underline{$22.933 \pm 1.189$} & \underline{$34.682 \pm 1.443$} \\
&                     & Transformer & $0.854 \pm 0.004$ & $29.560 \pm 1.234$ & $42.337 \pm 0.756$ \\
\hhline{|~|-|-|-|-|-|}
& \multirow{4}{*}{5M} & GINE        & $0.879 \pm 0.007$ & $26.733 \pm 0.811$ & $38.447 \pm 0.987$ \\
&                     & SchNet      & $0.861 \pm 0.008$ & $28.910 \pm 0.878$ & $41.205 \pm 1.132$ \\
&                     & DeBERTaV2   & \bm{$0.904 \pm 0.006$} & \bm{$22.700 \pm 1.150$} & \bm{$34.400 \pm 1.410$} \\
&                     & Transformer & $0.860 \pm 0.008$ & $28.910 \pm 1.137$ & $41.396 \pm 1.234$ \\
\hline

\multirow{8}{*}{$T_m$} 
& \multirow{4}{*}{2M} & GINE        & $0.660 \pm 0.027$ & $45.832 \pm 2.474$ & $63.087 \pm 2.744$ \\
&                     & SchNet      & $0.646 \pm 0.036$ & $47.079 \pm 2.048$ & $64.293 \pm 2.807$ \\
&                     & DeBERTaV2   & \underline{$0.754 \pm 0.022$} & \underline{$36.156 \pm 1.472$} & \underline{$53.588 \pm 2.608$} \\
&                     & Transformer & $0.617 \pm 0.017$ & $48.089 \pm 2.255$ & $66.937 \pm 2.420$ \\
\hhline{|~|-|-|-|-|-|}
& \multirow{4}{*}{5M} & GINE        & $0.697 \pm 0.022$ & $42.683 \pm 2.142$ & $59.475 \pm 2.443$ \\
&                     & SchNet      & $0.643 \pm 0.033$ & $47.282 \pm 1.836$ & $64.504 \pm 2.517$ \\
&                     & DeBERTaV2   & \bm{$0.756 \pm 0.021$} & \bm{$35.900 \pm 1.430$} & \bm{$53.300 \pm 2.560$} \\
&                     & Transformer & $0.638 \pm 0.025$ & $46.038 \pm 2.301$ & $65.041 \pm 2.326$ \\
\hline

\multirow{8}{*}{$T_d$} 
& \multirow{4}{*}{2M} & GINE        & $0.589 \pm 0.018$ & $51.326 \pm 1.586$ & $72.251 \pm 1.991$ \\
&                     & SchNet      & $0.583 \pm 0.023$ & $52.289 \pm 1.166$ & $72.778 \pm 2.220$ \\
&                     & DeBERTaV2   & \underline{$0.707 \pm 0.027$} & \underline{$40.473 \pm 2.273$} & \underline{$61.032 \pm 3.717$} \\
&                     & Transformer & $0.596 \pm 0.054$ & $49.899 \pm 4.635$ & $71.489 \pm 4.843$ \\
\hhline{|~|-|-|-|-|-|}
& \multirow{4}{*}{5M} & GINE        & $0.647 \pm 0.025$ & $47.133 \pm 2.386$ & $66.961 \pm 3.282$ \\
&                     & SchNet      & $0.576 \pm 0.016$ & $52.662 \pm 0.898$ & $73.400 \pm 1.836$ \\
&                     & DeBERTaV2   & \bm{$0.709 \pm 0.026$} & \bm{$40.200 \pm 2.240$} & \bm{$60.700 \pm 3.680$} \\
&                     & Transformer & $0.603 \pm 0.036$ & $50.374 \pm 3.255$ & $70.922 \pm 2.855$ \\
\hline

\end{tabular}}
\end{table*}

\textbf{Unimodal conditional generation (Table \ref{tab:generation_unimodal}):}
Generative evaluation separates decoder/representation feasibility from encoder-conditioned exploration. Across models and scales, validity is effectively saturated: the decoding pipeline and chemical-syntactic constraints dominate feasibility, and nearly all unimodal variants produce chemically valid candidates once trained. By contrast, novelty and structural diversity reveal clear modality-dependent differences. Sequence and fingerprint encoders produce smoother, more navigable conditional manifolds that support higher novelty and broader structural diversity under property constraints; this manifests as greater exploration away from training-set motifs. Graph- and geometry-based encoders tend toward more concentrated conditional outputs at smaller pretraining scale, consistent with mode concentration when representation coverage is limited; in some cases, increasing pretraining scale expands their conditional support and improves exploratory breadth, but these improvements are neither uniform nor guaranteed across all properties. Overall, the generative results show that decoder feasibility is solved by design, while the encoder modality and the extent of pretraining primarily determine whether conditional sampling explores meaningfully beyond memorized chemotypes.

\begin{table*}[h!]
\centering
\caption{\textbf{Effect of pretraining scale and modality-specific encoders on generative quality (validity, novelty, diversity).}
We report three generation metrics—Validity (percentage of chemically valid outputs), Novelty (percentage of generated polymers not present in the training set), and Diversity (internal structural diversity)—with values shown as mean $\pm$ standard deviation and higher values indicating better performance. Results are given for models pretrained on two corpus scales (2 million, 2M; and 5 million, 5M polymers), with each block of rows corresponding to one pretraining scale. The unimodal models compared are GINE (2D graph encoder), SchNet (3D conformation–aware encoder), DeBERTaV2 (PSMILES-based encoder), and a Transformer trained on engineered fingerprints. Reported values are the mean across five cross-validation runs and the uncertainty is the standard deviation across those runs. Boldface indicates the best-performing model for a given target and pretraining scale, and underlining indicates the second best. Overall, validity is uniformly high across models and scales, while DeBERTaV2 and the Transformer consistently achieve the top or near-top novelty and diversity; increasing pretraining scale often improves novelty and diversity for certain encoders, indicating that larger pretraining corpora can enhance generative breadth without sacrificing chemical validity.}
\label{tab:generation_unimodal}

\resizebox{\textwidth}{!}{
\begin{tabular}{|l|l|l|c|c|c|}
\hline
\textbf{Property} & \textbf{Corpora Scale} & \textbf{Model} & \textbf{Validity} (\%) $\uparrow$ & \textbf{Novelty} (\%) $\uparrow$ & \textbf{Diversity} $\uparrow$ \\
\hline

\multirow{8}{*}{$\rho$}
& \multirow{4}{*}{2M} & GINE        & 99.995 $\pm$ 0.006 & 41.684 $\pm$ 5.667 & 0.428 $\pm$ 0.045 \\
&                     & SchNet      & 99.995 $\pm$ 0.006 & 68.091 $\pm$ 11.961 & 0.640 $\pm$ 0.071 \\
&                     & DeBERTaV2   & 99.993 $\pm$ 0.006 & 83.445 $\pm$ 4.803 & 0.829 $\pm$ 0.022 \\
&                     & Transformer & 99.993 $\pm$ 0.010 & \textbf{83.795 $\pm$ 3.308} & \textbf{0.859 $\pm$ 0.015} \\
\hhline{|~|-|-|-|-|-|}
& \multirow{4}{*}{5M} & GINE        & \textbf{100.000 $\pm$ 0.000} & 52.854 $\pm$ 5.413 & 0.527 $\pm$ 0.038 \\
&                     & SchNet      & 99.995 $\pm$ 0.010 & 66.602 $\pm$ 9.212 & 0.612 $\pm$ 0.066 \\
&                     & DeBERTaV2   & \underline{99.998 $\pm$ 0.005} & \underline{83.710 $\pm$ 6.897} & 0.823 $\pm$ 0.011 \\
&                     & Transformer & \textbf{100.000 $\pm$ 0.000} & 83.627 $\pm$ 6.228 & \underline{0.843 $\pm$ 0.019} \\
\hline

\multirow{8}{*}{$T_g$}
& \multirow{4}{*}{2M} & GINE        & \underline{99.973 $\pm$ 0.027} & 30.984 $\pm$ 4.561 & 0.281 $\pm$ 0.040 \\
&                     & SchNet      & 99.841 $\pm$ 0.165 & 58.057 $\pm$ 16.408 & 0.479 $\pm$ 0.164 \\
&                     & DeBERTaV2   & 99.971 $\pm$ 0.031 & \underline{90.712 $\pm$ 2.660} & \underline{0.792 $\pm$ 0.039} \\
&                     & Transformer & 99.910 $\pm$ 0.095 & 79.976 $\pm$ 12.647 & 0.623 $\pm$ 0.191 \\
\hhline{|~|-|-|-|-|-|}
& \multirow{4}{*}{5M} & GINE        & 99.915 $\pm$ 0.056 & 32.461 $\pm$ 3.711 & 0.288 $\pm$ 0.036 \\
&                     & SchNet      & 99.795 $\pm$ 0.107 & 55.300 $\pm$ 10.445 & 0.464 $\pm$ 0.099 \\
&                     & DeBERTaV2   & \textbf{99.990 $\pm$ 0.012} & 89.748 $\pm$ 2.717 & 0.768 $\pm$ 0.025 \\
&                     & Transformer & 99.949 $\pm$ 0.057 & \textbf{92.379 $\pm$ 4.881} & \textbf{0.794 $\pm$ 0.028} \\
\hline

\multirow{8}{*}{$T_m$}
& \multirow{4}{*}{2M} & GINE        & 99.968 $\pm$ 0.029 & 36.585 $\pm$ 9.670 & 0.336 $\pm$ 0.076 \\
&                     & SchNet      & 99.983 $\pm$ 0.028 & 52.396 $\pm$ 8.592 & 0.448 $\pm$ 0.087 \\
&                     & DeBERTaV2   & \textbf{99.995 $\pm$ 0.010} & 80.709 $\pm$ 9.257 & 0.707 $\pm$ 0.129 \\
&                     & Transformer & 99.985 $\pm$ 0.014 & \textbf{89.201 $\pm$ 2.362} & \underline{0.824 $\pm$ 0.029} \\
\hhline{|~|-|-|-|-|-|}
& \multirow{4}{*}{5M} & GINE        & \textbf{99.995 $\pm$ 0.006} & 36.539 $\pm$ 9.689 & 0.343 $\pm$ 0.078 \\
&                     & SchNet      & 99.990 $\pm$ 0.014 & 54.757 $\pm$ 9.328 & 0.464 $\pm$ 0.068 \\
&                     & DeBERTaV2   & 99.968 $\pm$ 0.045 & 79.051 $\pm$ 15.676 & 0.684 $\pm$ 0.204 \\
&                     & Transformer & \underline{99.993 $\pm$ 0.010} & \underline{86.670 $\pm$ 5.394} & \textbf{0.833 $\pm$ 0.014} \\
\hline

\multirow{8}{*}{$T_d$}
& \multirow{4}{*}{2M} & GINE        & 99.912 $\pm$ 0.109 & 34.393 $\pm$ 1.021 & 0.314 $\pm$ 0.010 \\
&                     & SchNet      & \underline{99.978 $\pm$ 0.021} & 50.121 $\pm$ 3.534 & 0.442 $\pm$ 0.048 \\
&                     & DeBERTaV2   & 99.966 $\pm$ 0.046 & 84.132 $\pm$ 5.384 & 0.708 $\pm$ 0.069 \\
&                     & Transformer & 99.958 $\pm$ 0.040 & \underline{88.416 $\pm$ 3.891} & \underline{0.754 $\pm$ 0.103} \\
\hhline{|~|-|-|-|-|-|}
& \multirow{4}{*}{5M} & GINE        & \textbf{99.983 $\pm$ 0.012} & 31.946 $\pm$ 3.839 & 0.275 $\pm$ 0.095 \\
&                     & SchNet      & 99.932 $\pm$ 0.084 & 48.729 $\pm$ 3.092 & 0.403 $\pm$ 0.027 \\
&                     & DeBERTaV2   & 99.844 $\pm$ 0.182 & 78.835 $\pm$ 9.031 & 0.688 $\pm$ 0.116 \\
&                     & Transformer & 99.861 $\pm$ 0.169 & \textbf{89.910 $\pm$ 3.181} & \textbf{0.812 $\pm$ 0.050} \\
\hline
\end{tabular}}
\end{table*}

\textbf{Implication for multimodal fusion:}
Taken together, the unimodal findings motivate a fusion strategy that utilizes complementary modality strengths. Sequence encoders contribute reliable forward-prediction capacity linked to token-context capture of chemically informative motifs; structure-native encoders encode geometry- and graph-derived cues that can enhance mechanistic fidelity but require scale to reach their potential; and descriptor encoders provide a stable backbone that supports broad conditional exploration. These observations further strengthen PolyFusion’s cross-modal alignment: by co-embedding modalities that are individually strong for either prediction or conditional exploration, the fused representation inherits both the predictive reliability of structure-native encoders and the exploration capacity of descriptor- and sequence-native encoders, while mitigating the proxy noise intrinsic to repeat-unit graph and 3D constructions.

\subsection*{Supplementary Note 7 - Simulated Evaluation Cases for
PolyAgent}
\label{supp:note7}
\subsubsection*{Rationale and scope}
To evaluate PolyAgent in settings that resemble real polymer R\&D interactions (where questions mix mechanism, constraints, data-driven estimates, design, and experimental validation), we constructed a simulated case bank of 20 prompts spanning packaging, additive manufacturing, adhesives, cable jacketing, flexible electronics, composites, membranes, biomedical polymers, and recycling. Each case is formulated as a practitioner request with (i) a seed repeat-unit representation (PSMILES), (ii) explicit application constraints (e.g., halogen-free, process window, optical clarity), and (iii) multi-part tasks requiring both scientific reasoning and workflow planning. The goal is not to maximize prompt difficulty, but to probe whether the agent produces actionable, chemically plausible, and tool-grounded outputs under realistic ambiguity.

\subsubsection*{Case prompts}
\begin{itemize}
\item \textbf{Case 1 -- Humidity-sensitive recyclable film:}
I’m developing a recyclable, transparent monomaterial packaging film for ambient snacks. Constraints: halogen-free chemistry, plausible food-contact status, and compatibility with high-speed blown-film lines. My baseline repeat unit is a flexible aliphatic polyester:
\begin{quote}
\texttt{seed\_psmiles: [*]OCCOC(=O)CC(=O)OCCO[*]}
\end{quote}
Under humid conditions, the oxygen barrier collapses far more than I expected. Please analyze why this design is humidity-sensitive, focusing on:
Polarity distribution (where water “parks”),
Segmental mobility and plasticization,
Free volume changes and diffusion pathways,
And especially how Tg relative to use temperature (Tg \textasciitilde70--80 \textdegree C) can still lead to barrier failure via local mobility/free-volume swelling.
Then give a structure–property rationale for adding slightly more rigid aromatic or cyclic content to push Tg upward ($\ge$80 \textdegree C) while avoiding brittleness and limiting water uptake.
Finally, generate and rank \textasciitilde6 candidate repeat units likely to remain blow-film processable yet improve barrier via packing and crystallinity tuning, and outline a practical OTR/WVTR + tensile screening workflow (conditions, specimen handling, humidity preconditioning). 

\item \textbf{Case 2 -- Bio-based chilled-food film:}
I’m working on a bio-based, biodegradable film for refrigerated salad packs. I’m trying to tune lineal yield and stiffness primarily through bulk density, not by driving the material to very high crystallinity. Current backbone is a polar aliphatic polyester:
\begin{quote}
\texttt{seed\_psmiles: [*]OC(=O)CC(O)COC(=O)C[*]}
\end{quote}
First, reason about why this motif likely lands near \textasciitilde1.20 g/cm$^3$ (packing efficiency of short diacid/diol motifs, hydroxyl/ester polarity, etc.). Then, using property prediction, estimate density and explicitly tell me whether 1.20 g/cm$^3$ feels high or low (i.e., likely under- or over-estimate) and what features drive that direction.
Do not generate new candidates. Instead, discuss how incremental structural shifts---e.g., inserting aromatic rings vs. adding longer aliphatic spacers---would qualitatively move density up/down, and how those density moves would feed into flexural modulus and barrier under cold-chain conditions. Close by mapping these structural levers to at least 8 papers (recent) you would cite on density--stiffness--barrier relationships in biodegradable polyesters. (A short “why each paper matters” note would help.)

\item \textbf{Case 3 -- FDM filament tuning}
I’m evaluating a PLA-like filament for FDM printing (nozzle \textasciitilde200--210 \textdegree C). I want $T_m$ high enough for shape retention but not so high that printing becomes finicky. Baseline repeat unit:
\begin{quote}
\texttt{seed\_psmiles: [*]OC(=O)C(C)(COC(=O)O[*])c1ccccc1}
\end{quote}
My gut says this could melt around \textasciitilde190 \textdegree C, but I want a more defensible estimate. Please predict $T_m$ and tell me whether targeting \textasciitilde190 \textdegree C is a sensible operating point given typical FDM thermal profiles.
Then treat this as inverse design: generate \textasciitilde6 candidate repeat units that keep $T_m$ in \textasciitilde180--200 \textdegree C, but reduce brittleness / overly high crystallinity so parts don’t crack while cooling.
Finally, propose an experimental validation plan: DSC (including second heat), print trials for warpage/dimensional stability, and creep at service temperature. Also point out which structural “signals” in your candidates make you optimistic for printability (and what could go wrong).

\item \textbf{Case 4 -- Hot-melt adhesive layer}
I’m formulating a hot-melt adhesive layer in a multilayer packaging structure. I’m trying to balance shrinkage on cooling vs toughness. Current branched aliphatic polyester:
\begin{quote}
\texttt{seed\_psmiles: [*]CC(C)CCOC(=O)CCCOC(=O)CC(C)C[*]}
\end{quote}
It feels too bulky; I suspect the $T_m$ is on the low side---maybe around 60--90 \textdegree C---and I’d prefer inverse designs with a $T_m$ window near 80--110 \textdegree C to limit creep/print blocking without turning glassy/brittle.
Please reason through how branching, polarity, and crystallinity suppression are interacting, then estimate $T_m$ via property prediction (and tell me what the main uncertainty is).
Then inverse-design: generate several candidate repeat units likely to sit near $T_m$ \textasciitilde80--110 \textdegree C, compatible with standard hot-melt processing and halogen-free chemistry.
Wrap up with a quick screening protocol: dimensional stability after quench, T-peel (including temperature dependence), and DMA as a fast discriminator. 

\item \textbf{Case 5 -- Flexible cable jacketing}
I’m designing a flexible cable jacket for industrial equipment where continuous thermal exposure and short overloads matter. I’m exploring a phosphorus + aromatic motif:
\begin{quote}
\texttt{seed\_psmiles: [*]CCN(c1ccccc1)P(=O)(OCC)OCC[*]}
\end{quote}
I’m hoping this architecture pushes $T_d$ toward the high 300s \textdegree C while still allowing a workable processing window. Please analyze how P=O, aromatic groups, and backbone rigidity affect $T_d$, and provide a predicted decomposition onset.
Then generate a handful of candidate repeat units designed to keep flexibility but land in $T_d$ \textasciitilde360--400 \textdegree C, without halogens and without going to overly rigid imide-style structures that crack under bending.
End with a practical screening workflow: TGA (air vs N$_2$), UL-style flammability screen, and low-temperature bend/flex durability.

\item \textbf{Case 6 -- Flexible OLED substrate}
For a flexible OLED substrate, I need a transparent, colorless polymer that tolerates repeated bakes around 150--170 \textdegree C without permanent deformation. Current polyarylate-like unit:
\begin{quote}
\texttt{seed\_psmiles: [*]OC(=O)c1ccc(C(=O)O[*])c(C(F)(F)F)c1}
\end{quote}
Target is $T_g$ \textasciitilde190--210 \textdegree C (call 200 \textdegree C a practical goal). Please dissect how aromaticity, carbonyl density, and CF$_3$ affect $T_g$ and free volume, and then predict $T_g$.
Then inverse-design: generate and prioritize several repeat units that push $T_g$ toward \textasciitilde200 \textdegree C while maintaining optical clarity (avoid strongly absorbing chromophores) and reasonable film processability.
Finish with an experimental plan combining DSC, thermo-mechanical creep, and optical haze/clarity tests for fast down-selection.

\item \textbf{Case 7 -- Light-weight composite matrix}
For a glass-fiber reinforced composite in automotive body panels, I’m trying to reduce weight while keeping acceptable global stiffness (dominated by the glass fibers and fiber volume fraction) and improving the matrix-controlled thermal performance ($T_g$, HDT, creep). As a starting point, I consider a very aliphatic backbone with the following seed repeat unit:
\begin{quote}
\texttt{seed\_psmiles: [*]CC(C)CC(C)CC(C)C[*]}
\end{quote}
This highly saturated, branched hydrocarbon motif is expected to pack relatively loosely, giving a matrix density on the order of \textasciitilde1.00--1.05 g/cm$^3$ and a relatively low glass-transition temperature and matrix modulus due to its backbone flexibility. I am willing to accept a modest increase in density (up to roughly \textasciitilde1.10 g/cm$^3$) if it is accompanied by significantly improved thermal resistance (higher $T_g$/$T_m$) and stiffer matrix response, while still maintaining the lightweight benefit of the composite.

\item \textbf{Case 8 -- Peelable seal layer}
I’m tuning a peelable seal layer for lidding films. I want a melt transition that enables sealing in 115--130 \textdegree C without harming a stiffer core layer. Working repeat unit:
\begin{quote}
\texttt{seed\_psmiles: [*]CC(C)CCOC(=O)C(C)CCOC(=O)C[*]}
\end{quote}
Please predict $T_m$ and state clearly whether it likely falls inside, below, or above the 115--130 \textdegree C sealing window.
Don’t generate new repeat units. Instead, explain how subtle changes---branch length, comonomer insertion, tacticity---shift $T_m$ and seal behavior, and connect these levers to hot-tack, seal plateau, and abuse resistance. Please cite \textasciitilde10 recent packaging papers linking structure/$T_m$ to seal performance, and briefly note what each paper supports. 

\item \textbf{Case 9 -- CO$_2$-selective membrane}
For a glassy gas-separation membrane (CO$_2$/CH$_4$), I’m evaluating a rigid aromatic polycarbonate-like unit:
\begin{quote}
\texttt{seed\_psmiles: [*]OC(=O)Oc1ccc(C(C)(C)c2ccc(OC(=O)O[*])cc2)cc1}
\end{quote}
Conceptually I expect relatively high free volume and a specific volume maybe around 0.80--0.85 cm$^3$/g. Please estimate specific volume quantitatively and explain what features dominate the prediction.
No new structures, but discuss how adding contorted/spiro moieties would further raise free volume and how that typically trades permeability vs selectivity for CO$_2$/CH$_4$, assuming $T_g$ stays well above 100 \textdegree C.
Anchor the discussion in 2015--2025 membrane literature with at least 18 sources relating free volume/specific volume and gas transport performance.

\item \textbf{Case 10 -- Proton-conducting membrane}
In a sulfonated aromatic polymer concept for PEMs, I’m looking at a simplified motif:
\begin{quote}
\texttt{seed\_psmiles: [*]c1ccc(cc1)S(=O)2OCCOC2[*]}
\end{quote}
I’m worried whether aromatic stability plus sulfone-like functionality can keep $T_d \ge 300$ \textdegree C. Please reason through likely thermal behavior, predict $T_d$, and compare to the 300 \textdegree C benchmark.
No alternative generation. Discuss how adding fluorinated groups or additional sulfonic-acid-like groups would shift $T_d$, $T_g$, and mechanical robustness, and relate that to membrane lifetime under harsh oxidative fuel-cell conditions.
Wrap with \textasciitilde11 literature references on how $T_d$ and backbone architecture constrain PEM operating windows.

\item \textbf{Case 11 -- Battery separator}
For a lithium-ion battery separator, I want a polyolefin-like matrix that, when processed into a stretched microporous film, begins shutting down porosity near 140--150 \textdegree C but avoids catastrophic shrinkage until just below $T_m$ \textasciitilde160 \textdegree C. Minimal model repeat unit:
\begin{quote}
\texttt{seed\_psmiles: [*]CCCCCCCCC[*]}
\end{quote}
Discuss how symmetry/chain length affects crystallinity and $T_m$; predict $T_m$ and compare to 160 \textdegree C as the design anchor.
Then inverse-design: generate candidate repeat units (light branching, mild comonomer disruption, etc.) that keep $T_m$ in \textasciitilde150--165 \textdegree C while managing shrinkage and preserving electrochemical stability.
Close with a pragmatic validation plan: DSC, shrinkage under load (with defined ramp), and electrolyte wetting/contact angle/uptake.

\item \textbf{Case 12 -- Stretchable encapsulant}
I’m choosing an ultra-soft encapsulant for stretchable electronics that must stay rubbery down to at least $-10$ \textdegree C. Siloxane-like unit:
\begin{quote}
\texttt{seed\_psmiles: [*]O[Si](C)(C)O[Si](C)(C)O[*]}
\end{quote}
Please explain how the Si--O backbone and methyl substituents drive low $T_g$ and higher free volume, then predict $T_g$ and compare to a design target of \textasciitilde$-30$ \textdegree C.
Then inverse-design: generate candidate repeat units ($T_g < -20$ \textdegree C) that improve adhesion to flexible PCB substrates or elastomeric overmolds while keeping softness.
End with a rapid screening plan: DMA down to low temp, cyclic strain durability, and adhesion (peel/lap shear) in relevant humidity/aging.

\item \textbf{Case 13 -- Dense barrier bottle resin}
For a refillable “premium feel” small-format bottle, I want an aromatic polyester with high barrier and higher density. Candidate unit:
\begin{quote}
\texttt{seed\_psmiles: [*]OC(=O)c1ccc(cc1)C(=O)OCCOC(=O)c2ccc(cc2)C(=O)O[*]}
\end{quote}
Please analyze how aromatic diacid/diol choices impact density and barrier, then predict density and compare explicitly to $\rho = 1.30$ g/cm$^3$.
Then inverse-design: generate candidate repeat units that stay in 1.25--1.35 g/cm$^3$ while preserving melt-processability typical of bottle resins and halogen-free chemistry.
Close with an experimental plan: density measurement (method), gas barrier tests, and impact resistance ranking. 

\item \textbf{Case 14 -- Structural foam core}
For a polymer foam core in lightweight sandwich panels, I want high expansion to reduce density but must keep reasonable stiffness and heat resistance. Starting repeat unit:
\begin{quote}
\texttt{seed\_psmiles: []OCC(COC(=O)CC(CO)COC(=O)CC)CO[]}
\end{quote}
I suspect $T_g$ is modest (near 0--30 \textdegree C) due to branching and flexibility, but I want $T_g$ closer to 60--90 \textdegree C (for heat resistance) while keeping foamability and processing latitude.
Please reason about how branching, flexibility, and crystallinity influence $T_g$ and foam morphology, then predict $T_g$.
Then inverse-design: generate repeat units likely to push $T_g$ toward \textasciitilde60--90 \textdegree C while retaining stiffness in the cell walls and reasonable processing.
Finish with fast screening: foam expansion, compressive modulus, and thermal softening (with test conditions). 

\item \textbf{Case 15 -- High-temperature composite resin}
For an aerospace carbon-fiber composite matrix, I need a resin that tolerates excursions toward 350--400 \textdegree C without major mass loss. Simplified imide-like motif:
\begin{quote}
\texttt{seed\_psmiles: [*]N(C(=O)c1ccc(cc1)C(=O)N)c2ccc(cc2)C(=O)O[*]}
\end{quote}
Please analyze expected behavior under high-temperature oxidative conditions and predict $T_d$; compare to an aspirational $T_d$ \textasciitilde450 \textdegree C.
Then inverse-design: generate/prioritize candidate repeat units in $T_d$ \textasciitilde430--470 \textdegree C that are still processable (melt or RTM compatible) and compatible with carbon fibers.
Close with an experimental matrix: TGA (air/N$_2$), rheology vs time/temperature, and open-hole compression after thermal aging.

\item \textbf{Case 16 -- Biodegradable microspheres}
For biodegradable polyester microspheres in long-acting injectables, I’m trying to tune release kinetics by setting $T_g$ relative to 37 \textdegree C. Candidate unit:
\begin{quote}
\texttt{seed\_psmiles: [*]OC(=O)CH(CH3)OC(=O)CH2CH2OC(=O)CH(CH3)O[*]}
\end{quote}
I worry $T_g$ might be “too close” to body temperature. Please reason through how side groups and polarity affect mobility, predict $T_g$, and clearly state whether $T_g$ sits above or below 37 \textdegree C.
No new structures. Discuss how comonomer ratio shifts or molecular weight changes would move $T_g$ and the balance between diffusion- vs erosion-controlled release.
Tie to \textasciitilde9 recent papers on biodegradable microsphere design (and note which aspect each paper supports).

\item \textbf{Case 17 -- Biodegradable mulch film}
For mulch films that must survive installation but degrade after a season, I want $T_m$ high enough to resist blocking in warm storage yet low enough for standard blown-film extrusion. Representative repeat unit:
\begin{quote}
\texttt{seed\_psmiles: [*]OC(=O)CC(OC(=O)CCOC(=O)CC)CO[*]}
\end{quote}
Please predict $T_m$ and compare it to a working target near 110 \textdegree C (tell me if that target is aggressive or conservative).
Then generate candidate repeat units that keep $T_m$ in \textasciitilde100--120 \textdegree C but improve toughness and tear resistance (qualitatively) without destroying processability.
Finish with a realistic screening plan: DSC, tensile + tear, and controlled outdoor exposure. Tie reasoning to at least 14 recent mulch-film studies. 

\item \textbf{Case 18 -- Medical device handle}
For injection-molded medical device handles, I want a polymer that feels substantial (“heft”) but remains chemically resistant and sterilizable. Mixed aromatic--aliphatic ester motif:
\begin{quote}
\texttt{seed\_psmiles: [*]CCOC(=O)c1ccc(cc1)CCOC(=O)CCc2ccc(cc2)O[*]}
\end{quote}
Please predict density and state clearly whether it’s closer to 1.1 or 1.3 g/cm$^3$. Then discuss how swapping some aliphatic segments for heavier/more polar groups (no halogens) would shift density, modulus, and perceived heft.
Do not generate new PSMILES. Instead, connect the structure--density--haptics discussion to \textasciitilde7 literature examples relevant to handheld medical or consumer devices. 

\item \textbf{Case 19 -- Optical fiber primary coating}
For an optical fiber primary coating, I need a very soft, low-modulus material where $T_g$ minimizes microbending losses. Candidate urethane-acrylate-like backbone:
\begin{quote}
\texttt{seed\_psmiles: [*]OCCOC(=O)NCCOC(=O)OCCOCCO[*]}
\end{quote}
Please explain how ether flexibility and urethane polarity affect $T_g$, then estimate position $T_g$ qualitatively vs room temperature.
No new repeat units. Discuss how adding bulky side groups vs inserting more flexible spacers would alter $T_g$, stress relaxation, and long-term optical performance.
Tie to at least 8 published studies on $T_g$/microbending in fiber coatings.

\item \textbf{Case 20 -- Compatibilizer for recycled PE/PP}
For compatibilizing mixed-stream recycled PE/PP in durable goods, I need a soft, polar compatibilizer that survives elevated-temperature reprocessing while staying rubbery at room temperature. Concept repeat unit:
\begin{quote}
\texttt{seed\_psmiles: [*]CC(COOC(=O)CCCC)CCOC(=O)CCC[*]}
\end{quote}
Please reason how this structure will behave across multiple melt cycles, then predict $T_d$ quantitatively and comment on $T_g$ qualitatively (relative to 25 \textdegree C).
Then inverse-design: generate candidate repeat units that keep $T_d$ near \textasciitilde350 \textdegree C while maintaining $T_g$ below room temperature, and explain how their polarity/architecture should improve PE--PP interfacial adhesion.
Close with a concise screening plan: TGA, rheology over multiple reprocessing cycles, and mechanical testing of PE/PP blend plaques. Anchor with at least 15 recent studies on compatibilizers and recycling.
\end{itemize}

\subsection*{Supplementary Note 8 -- Scoring Sheet Template for PolyAgent Evaluation}
\label{supp:note8}
Moreover, to ensure reproducible and fair human evaluation across the 20 simulated cases, we provide a sample scoring sheet (Table \ref{tab:ss}) that matches the evaluation rubric and fields used throughout the PolyAgent evaluation. 

\FloatBarrier

\begin{table*}[h!]
\centering
\caption{\textbf{Scoring sheet used for human evaluation of PolyAgent and LLM-only baselines.} Template used by human annotators to score the tool-augmented controller (PolyAgent) and two open-weight baselines (Mixtral-8$\times$22B and Llama-3.1-8B) on a common suite of 20 simulated polymer-design cases (Supplementary~\ref{supp:note7}). Rows correspond to the five rubric-defined metrics \cite{ferber2025development} — Tool Use, Completeness, Correctness, Helpfulness, and Citation Accuracy—each scored on a discrete 0–10 scale. For every metric and system, annotators record an integer score in the corresponding “Score (0–10)” cell and provide free-text justification in the “Notes / Evidence” cell, including references to specific tool calls, numerical estimates, constraint checks, and citations observed in the model’s response. This structure enforces side-by-side, metric-aligned comparison of PolyAgent against unaided LLMs while preserving a traceable rationale for each score, enabling aggregation across five independent evaluators and post hoc inspection of failure modes and disagreement.}
\label{tab:ss}

\resizebox{\linewidth}{!}{%
\begin{tabular}{| l | l | l | l | l | l | l |}
\hline
\textbf{Metric} &
\makecell[l]{\textbf{PolyAgent}\\\textbf{(Our system)}\\\textbf{Score (0--10)}} &
\makecell[l]{\textbf{PolyAgent}\\\textbf{(Our system)}\\\textbf{Notes / Evidence}} &
\makecell[l]{\textbf{Mixtral-8$\times$22B}\\\textbf{Score (0--10)}} &
\makecell[l]{\textbf{Mixtral-8$\times$22B}\\\textbf{Notes / Evidence}} &
\makecell[l]{\textbf{Llama-3.1-8B}\\\textbf{Score (0--10)}} &
\makecell[l]{\textbf{Llama-3.1-8B}\\\textbf{Notes / Evidence}} \\
\hline
\textbf{Tool Use} & &   & &   & &   \\
\hline
\textbf{Completeness} & &   & & & & \\
\hline
\textbf{Correctness} & & & & & &   \\
\hline
\textbf{Helpfulness} & & & & & & \\
\hline
\textbf{Citation Accuracy} & & & & & & \\
\hline
\end{tabular}%
}
\end{table*}


\bibliography{bibliography}